\documentclass{article}

\usepackage{color,xcolor}
\usepackage{epsfig}
\usepackage{graphicx}
\usepackage{algorithm,algorithmic}

\usepackage{adjustbox}
\usepackage{array}
\usepackage{booktabs}
\usepackage{colortbl}
\usepackage{float,wrapfig}
\usepackage{framed}
\usepackage{hhline}
\usepackage{multirow}
\usepackage{subcaption} 
\usepackage[font=small]{caption}
\usepackage[percent]{overpic}

\usepackage{amsmath,amsfonts,amssymb}
\usepackage{amsthm} 
\usepackage{bm}
\usepackage{nicefrac}
\usepackage{microtype}
\usepackage{contour}
\usepackage{courier}

\usepackage{changepage}
\usepackage{extramarks}
\usepackage{fancyhdr}
\usepackage{lastpage}
\usepackage{setspace}
\usepackage{soul}
\usepackage{xspace}
\usepackage{cuted}
\usepackage{fancybox}
\usepackage{afterpage}

\usepackage{url}
\usepackage{quoting}
\usepackage{epigraph}

\usepackage{enumerate}
\usepackage{paralist,tabularx}
\usepackage{comment}
\usepackage{pdfpages}

\usepackage{cite}
\usepackage{subcaption}
\usepackage{paralist}
\usepackage{float}
\usepackage{makecell}
\usepackage{graphicx}
\usepackage{amsmath}
\usepackage{bm}
\usepackage{amssymb}
\usepackage{array}

\usepackage[final]{corl_2020} 

\title{Hardware as Policy: Mechanical and Computational Co-Optimization using Deep Reinforcement Learning}

\author{
  Tianjian Chen\thanks{Authors have contributed equally.},~ Zhanpeng He\footnotemark[1],~ Matei Ciocarlie \\
  Columbia University \\
  \texttt{\{tianjian.chen, zhanpeng.he, matei.ciocarlie\}@columbia.edu} \\
}
  
\begin{document}
\maketitle

\vspace{-7mm}

\begin{abstract}
Deep Reinforcement Learning (RL) has shown great success in learning complex control policies for a variety of applications in robotics. However, in most such cases, the hardware of the robot has been considered immutable, modeled as part of the environment. In this study, we explore the problem of learning hardware and control parameters together in a unified RL framework. To achieve this, we propose to model the robot body as a ``hardware policy'', analogous to and optimized jointly with its computational counterpart. We show that, by modeling such hardware policies as auto-differentiable computational graphs, the ensuing optimization problem can be solved efficiently by gradient-based algorithms from the Policy Optimization family. We present two such design examples: a toy mass-spring problem, and a real-world problem of designing an underactuated hand. We compare our method against traditional co-optimization approaches, and also demonstrate its effectiveness by building a physical prototype based on the learned hardware parameters. Videos and more details are available at \url{https://roamlab.github.io/hwasp/} .
\end{abstract}

\vspace{-2mm}
\section{Introduction}
\vspace{-2mm}
\label{sec:intro}
Human ``intelligence'' resides in both the brain and the body: we can develop complex motor skills, and the mechanical properties of our bones and muscles are also adapted to our daily tasks. Numerous motor skills exhibit this phenomenon, from running (where the stiffness of the Achilles tendon has been shown to maximize locomotion efficiency~\cite{lichtwark2007achilles}) to grasping (where coordination patterns between finger joints emerge from both synergistic muscle control and mechanical coupling of joints~\cite{santello2013neural}). Mechanical adaptation and motor skill improvement can happen simultaneously, both over an individual's lifetime (e.g. \cite{hammami2016associations}) and at evolutionary time scales --- for example, it has been suggested that, as early hominids practiced throwing and clubbing, hand morphology also changed accordingly, e.g. the thumb got longer to provide better opposition~\cite{young2003evolution}.

In robotics, the idea of jointly designing/optimizing mechanical and computational components has a long track record with remarkable advances, exploiting the fact that the morphology, transmissions, and control policies are tightly connected by the laws of physics and co-determine robot behavior. If the control policy and dynamics can both be modeled analytically, traditional optimization can derive the desired values for hardware and policy parameters. When such an approach is not feasible (e.g. due to complex policies or dynamics), evolutionary computation has been used instead. However, these methods still have difficulty learning sophisticated motor skills in complex environments (e.g. with partially observable states, dynamics with transient contacts), or are sample-inefficient. 

\begin{figure*}
    \begin{subfigure}{\linewidth}
        \centering
        \includegraphics[width=.8\linewidth]{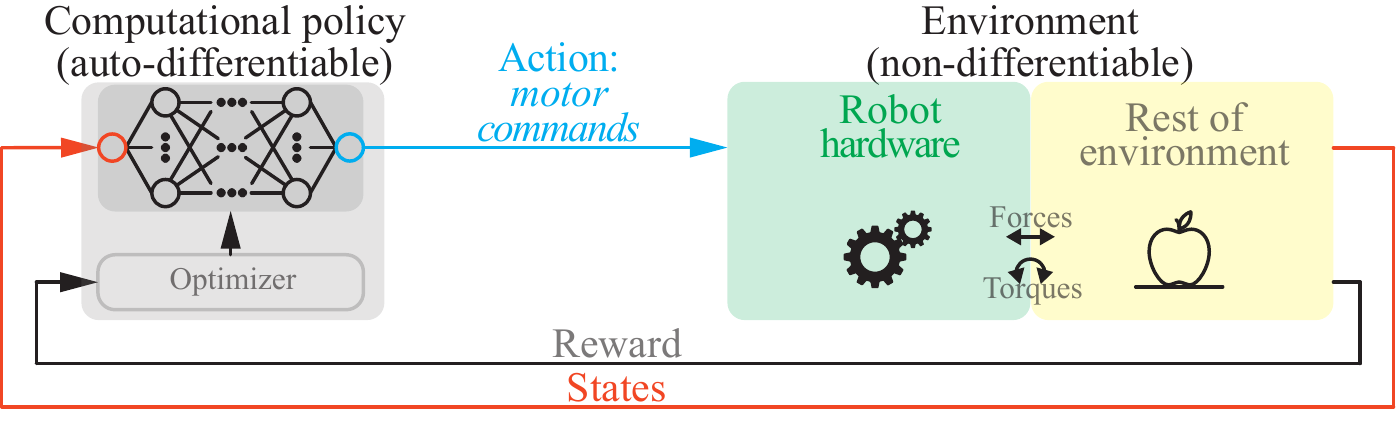}
        \caption{Traditional perspective --- Reinforcement Learning with a purely computational policy}
        \label{subfig:traditional_rl}
        \vspace{2mm}
    \end{subfigure}
    \newline
    \begin{subfigure}{\linewidth}
        \centering
        \includegraphics[width=.8\linewidth]{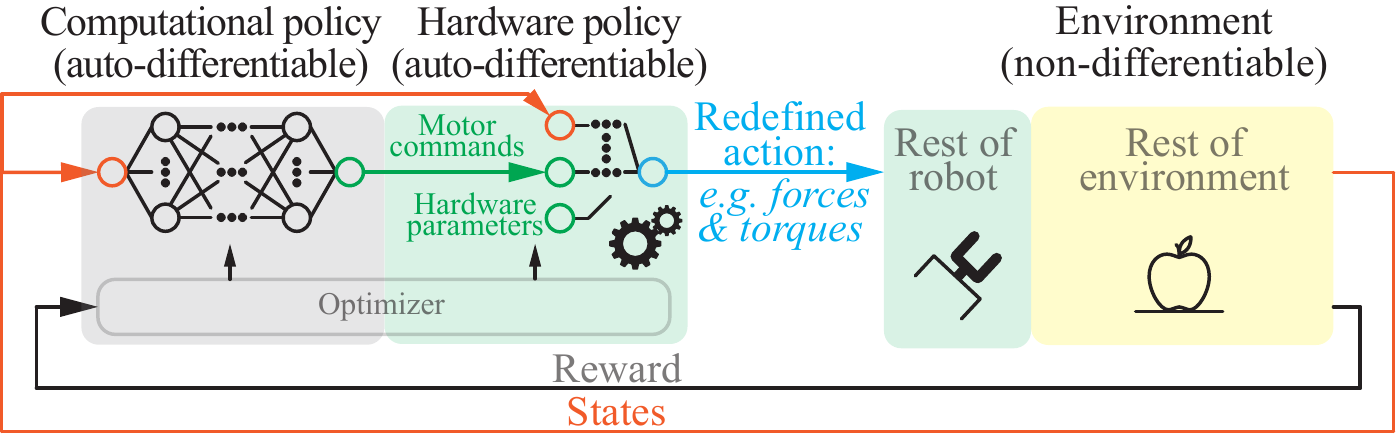}
        \caption{Hardware as Policy --- computational graph implementation}
        \label{subfig:hw_as_policy}
    \end{subfigure}

    \caption{Hardware as Policy overview. From the traditional perspective (a), all robot hardware is part of the simulated environment. In the proposed method (b), aspects of robot hardware are formulated as a ``hardware policy'' implemented as a computational graph, then optimized jointly with the computational policy. 
    }
    \label{fig:traditional_vs_proposed}
    \vspace{-6mm}
\end{figure*}

In contrast, recent advances in Deep Reinforcement Learning (Deep RL) have shown great potential for learning difficult motor skills despite having only partial information of complex, unstructured environments (e.g. \cite{rajeswaran2017learning, andrychowicz2018learning, haarnoja2018learning}). Traditionally, the output of a Deep RL policy in robotics consists of motor commands, and the robot hardware converts these motor commands to effects on the external world (usually through forces and/or torques). In this conventional RL perspective, robot hardware is considered given and immutable, essentially treated as part of the environment (Fig~\ref{fig:traditional_vs_proposed}a). 

In this work we bring a different perspective for hardware in RL. Consider the concrete example of an underactuated robot hand. Motor forces are converted into joint torques by a transmission mechanism ( gears, tendons or linkages). Through careful design of the hardware parameters, such a transmission can provide desired grasping behavior for a wide range of objects (e.g. \cite{birglen2004kinetostatic,odhner2014compliant}). Such a transmission is conceptually akin to a policy, mapping an input (motor forces) to an output (joint torques) with carefully tuned parameters leading to beneficial effects for overall performance.

Can we leverage the power of Deep RL for co-optimization of the computational and mechanical components of a robot? High-fidelity physics simulation and effective sim-to-real transfer (where a policy is trained on a physics simulator and only then deployed on real hardware~\cite{tobin2017domain}) provide such an opportunity, since they allows modifications of design parameters during training without incurring the prohibitive cost of re-building hardware. A straightforward option to optimize hardware parameters during training is to treat them as hyperparameters of the RL algorithm. However, this approach usually carries a massive computational cost.

In this study, we propose to consider \textit{hardware as policy}, optimized jointly with the traditional computational policy. As is well known, a model-free  Policy Optimization (e.g. \cite{schulman2015trust, schulman2017proximal}) or Actor-critic (e.g. \cite{lillicrap2015continuous}) algorithm can train using an auto-differentiable agent/policy and a non-differentiable black-box environment. The core idea we propose is to move part of the robot hardware from the non-differentiable environment into the auto-differentiable agent/policy (Fig~\ref{fig:traditional_vs_proposed}b). 
In this way, hardware parameters\footnote{This paper primarily focuses on the mechanical aspect of hardware, we use the terms ``hardware'' and ``mechanics/mechanical'' interchangeably. However, we believe that, in the future, the proposed idea could be extended to electrical or sensorial aspects of a physical device.} become parameters in the policy graph, analogous to the neural network weights and biases. Therefore, the optimization of hardware parameters can be directly incorporated into the existing RL framework, and can use existing learning algorithms with changes to the computational graphs as we will describe here.  We summarize our major contribution as follows: 
\begin{compactitem}
    \item To the best of our knowledge, we are the first to express hardware and computation as a unified RL policy which allows the optimization algorithm to propagate gradients of the actions w.r.t both hardware and computational parameters simultaneously.
    \item Via case studies comprising both a toy problem and a real-world design challenge, we show that such gradient-based methods are superior to hyperparameter tuning as well as gradient-free evolutionary strategies for hardware-policy co-optimization.
    \item To the best of our knowledge, we are the first to build a physical prototype and deploy the trained policy on it to validate a Deep RL-based hardware-policy co-optimization approach.
\end{compactitem}

\vspace{-2mm}
\section{Related Work}
\label{sec:related_work}
\vspace{-2mm}


A first category of related work comprises studies using analytical dynamics and classical control~\cite{park1994concurrent}, a number of which optimized mechanical and control or planning parameters for legged locomotors~\cite{paul2001road,geijtenbeek2013flexible,ha2018computational}. All studies above require an analytical model of the complete mechanical-control system, which is non-trivial in complex problems. 
More recent work uses classical control but evaluates and iterates on real hardware~\cite{liao2019data}, applying Bayesian Optimization to micro robots. However, the goal is different from ours: this study aims to decrease the number of real-world design evaluations, which is avoided in our work by training in simulation and performing sim-to-real transfer.

Evolutionary computation provides another way to approach this problem. This research path originated from studies on the evolution of artificial creatures \cite{sims1994evolving}, where the morphology and the neural systems are both encoded as graphs and generated using genetic algorithms. Lipson and Pollack \cite{lipson2000automatic} introduced the automatic lifeform design technique using bars, joints, and actuators as building blocks of the morphology, with neurons attached to them as controllers. A series of works from Cheney et al. \cite{cheney2014unshackling, cheney2016topological} studied the morphology-computation co-evolution of cellular automata, in the context of locomotion. Nygaard et al. \cite{nygaard2018real} presented a method that optimizes the morphology and control of quadruped robot using real-world evaluation of the robot.
Evolutionary strategies, which are gradient-free, have significant promise, but also exhibit high computational complexity and data-inefficiency compared to recent gradient-based optimization methods.

The recent influx of reinforcement learning provides a new perspective on the co-optimization problem.
Ha \cite{ha2018reinforcement} augmented the REINFORCE algorithm with rewards calculated using the mechanical parameters.
Schaff et al. \cite{schaff2019jointly} proposed a joint learning method to construct and control the agent, which models both design and control in a stochastic fashion and optimizes them via a variation of Proximal Policy Optimization (PPO).
Vermeer et al. \cite{vermeer2018kinematic} perform two-dimensional linkage mechanism synthesis using a Decision-Tree-based mechanism representation fused with RL. 
Luck et al. \cite{luck2019coadapt} presented a method for data-efficient co-adaptation of morphology and behaviors based on Soft Actor-Critic (SAC), leveraging previous information to estimate the performance of new candidates. 
In all the studies above, hardware parameters are still optimized iteratively and separately from the computational policies, whereas we aim to optimize both together in a unified framework. In addition, none of these works show physical prototypes based on the co-optimized agent.

Recent work on auto-differentiable physics~\cite{de2018end,degrave2019differentiable,hu2019chainqueen, hu2019difftaichi} is also relevant to us, as we rely on modeling (part of) the robot hardware as an auto-differentiable computational graph. We hope to make use of advances in general differentiable physics simulation in future iterations of our method.

\vspace{-2mm}
\section{Preliminaries}
\vspace{-2mm}
We start from a standard RL formulation, where the problem of optimizing an agent to perform a certain task can be modeled as a Markov Decision Process (MDP), represented by a tuple $(\mathcal{S}, \mathcal{A}, \mathcal{F}, \mathcal{R})$, where $\mathcal{S}$ is state space, $\mathcal{A}$ is the action space, $\mathcal{R}(\bm{s}, \bm{a})$ is the reward function, and $\mathcal{F}(\bm{s}' | \bm{s}, \bm{a})$ is the state transition model ($\bm{s}, \bm{s}' \in \mathcal{S}$, $\bm{s}'$ is for the next time step, $\bm{a} \in \mathcal{A}$). 
Behavior is determined by a computational control policy $\pi_{\bm{\theta}}^{comp}(\bm{a}|\bm{s})$, where $\bm{\theta}$ represents the parameters of the policy. Usually, $\pi_{\bm{\theta}}^{comp}$ is represented as a deep neural network, with $\bm{\theta}$ consisting of the network's weights and biases. The goal of learning is to find the values of the policy parameters that maximize the expected return $\mathbb{E}[\sum_{t=0}^T\mathcal{R}(\bm{s}_t,\bm{a}_t)]$ where $T$ is the length of episode.

We start from the observation that, in robotics, in addition to the parameters $\bm{\theta}$ of the computational policy, the design parameters of the hardware itself, denoted here by $\bm{\phi}$, play an equally important role for task outcomes. In particular, hardware parameters $\bm{\phi}$ help determine the output (the effect on the outside world) that is produced by a given input to the hardware (motor commands). This is analogous to how computational parameters $\bm{\theta}$ help determine the output of the computational policy (action $\bm{a}$) that is produced by a given input (state or observations $\bm{s}$). 

Even though this analogy exists, traditionally, these two classes of parameters have been treated very differently in RL. Computational parameters can be optimized via gradient-based methods: for example, in Policy Optimization methods such as Trust Region Policy Optimization (TRPO) \cite{schulman2015trust} and Proximal Policy Optimization (PPO) \cite{schulman2017proximal}, the parameters of the computational policy are optimized by computing and following the policy gradient: 
$\bm{{g}} = \mathbb{E}_{\tau\sim\pi_{\bm{\theta}}^{comp}}[\sum_{t=0}^T\nabla_{\bm{\theta}}\log\pi_{\bm{\theta}}^{comp}(\bm{a}_{t}|\bm{s}_{t}){A}_{t}(\bm{s}_t, \bm{a}_t)]$ ($A_t$ is the advantage function). 
In contrast, hardware is generally considered immutable, and modeled as part of the environment. Formally, this means that hardware parameters $\bm{\phi}$ are considered as parameters of the transition function $\mathcal{F}=\mathcal{F}_{\bm{\phi}}(\bm{s}' | \bm{s}, \bm{a})$ instead of the policy. This is the concept illustrated in Fig.~\ref{subfig:traditional_rl}. 
Such a formulation is grounded in the most general RL framework, where $\mathcal{F}$ is not modeled analytically, but only observed by execution on real hardware. In such a case, changing $\bm{\phi}$ can only be done by building a new prototype, which is generally impractical.

However, in recent years, the robotics community has made great advances in training via a computational model of the transition function $\mathcal{F}$, often referred to as a physics simulator (e.g. \cite{todorov2012mujoco}). The main drivers have been the need to train using many more samples than possible with real hardware, and ensure safety during training. Recent results have indeed shown that it is often possible to train exclusively using an imperfect analytical model of $\mathcal{F}$, and then transfer to the real world~\cite{tobin2017domain}.

In our context, training with such physics simulator opens new possibilities for hardware design: we can change the hardware parameters $\bm{\phi}$ and test different hardware configurations on-the-fly inside the simulator, without incurring the cost of re-building a prototype.

\vspace{-2mm}
\section{Hardware as Policy}
\vspace{-2mm}
\label{sec:method}
The Hardware as Policy method (HWasP) proposed here largely aims to perform a similar optimization for hardware parameters as we do for computational policy parameters, i.e. by computing and following the gradient of action probabilities w.r.t such parameters. 

The core of the HWasP method is to model the effects of the robot hardware we aim to optimize separately from the rest of the environment. We refer to this component as a “hardware policy”, and denote it via
$\pi_{\bm{\phi}}^{hw}(\bm{a}^{new}|\bm{s}, \bm{a})$. 
The input to the hardware policy consists of the action produced by the computational policy (i.e. a motor command) and other components of the state; the output is in a redefined action space $\mathcal{A}^{new}$ further discussed below. 

In the traditional formulation outlined so far, the ``hardware policy'' and its parameters $\bm{\phi}$ are included in the transition function $\mathcal{F}_{\bm{\phi}}$.
With HWasP, $\pi_{\bm{\phi}}^{hw}$ becomes part of the agent. The new overall policy 
$\pi_{\bm{\theta}, \bm{\phi}} = \pi_{\bm{\phi}}^{hw}(\bm{a}^{new} | \bm{s}, \bm{a})\pi_{\bm{\theta}}^{comp}(\bm{a} | \bm{s})$ comprises the composition of both computational and mechanical policies, while the new transition probability $\mathcal{F}^{new}  = \mathcal{F}^{new}(\bm{s}' | \bm{s}, \bm{a}^{new})$ encapsulates the rest of the environment. In other words, we have split the simulation of the environment: one part consists of the mechanical policy, now considered part of the agent, while the other simulates the rest of the robot, and the external environment. The reward function, $ \mathcal{R}(\bm{s}, \bm{a})$, is redefined to be associated with the new action space: $\mathcal{R}^{new}(\bm{s}, \bm{a}^{new})$.
Once this modification is performed, we run the original Policy Optimization algorithm on the new tuple $(\mathcal{S}, \mathcal{A}^{new}, \mathcal{F}^{new}, \mathcal{R}^{new})$ as redefined above. However, in order for this to be feasible, two conditions have to be met:

\textit{Condition 1:} The redefined action vector $\bm{a}^{new}$ must encapsulate the interactions between the mechanical policy and the rest of the environment. In other words, this new action interface must comprise all the ways in which the hardware we are optimizing effects change on the rest of the environment. Furthermore, the redefined action vector must be low-dimensional enough to allow for efficient optimization. Such an interface is problem-specific. Forces / torques between the robot and the environment make good candidates, as we will exemplify in the following sections. 


\textit{Condition 2:} To use Policy Optimization algorithms, we need to efficiently compute the gradient of the redefined action probability w.r.t. hardware parameters. We further discuss this condition next.

\textbf{Computational Graph Implementation (HWasP).} In order to meet Condition 2 above, \textit{we propose to simulate the part of hardware we care to optimize as a computational graph}. In this way, gradients can be computed by auto-differentiation and can flow or back-propagate through the hardware policy. Similar to the computational policy, the gradient of log-likelihood of actions w.r.t mechanical parameters $\bm{\phi}$ can be computed as $\nabla_{\bm{\phi}}\log\pi_{\bm{\phi}}^{hw}(\bm{a}^{new}|\bm{a}, \bm{s})$.

Critically, since the computational policy is also expressed as a computational graph, the gradient can back-propagate through both the hardware and computational policies, and the hardware and computational parameters are jointly optimized. This general idea is illustrated in Fig.~\ref{subfig:hw_as_policy}.

However, this approach is predicated on being able to simulate the effects of the hardware being optimized as a computational graph. Once again, the exact form of this simulation is problem-specific, and can be considered as a key part of the algorithm. In the next sections, we illustrate how this can be done both for a toy problem, and for a real-world design problem, and regard these implementations as an intrinsic part of the contribution of this work. 

\begin{figure*}
        \centering
        \includegraphics[width=.8\linewidth]{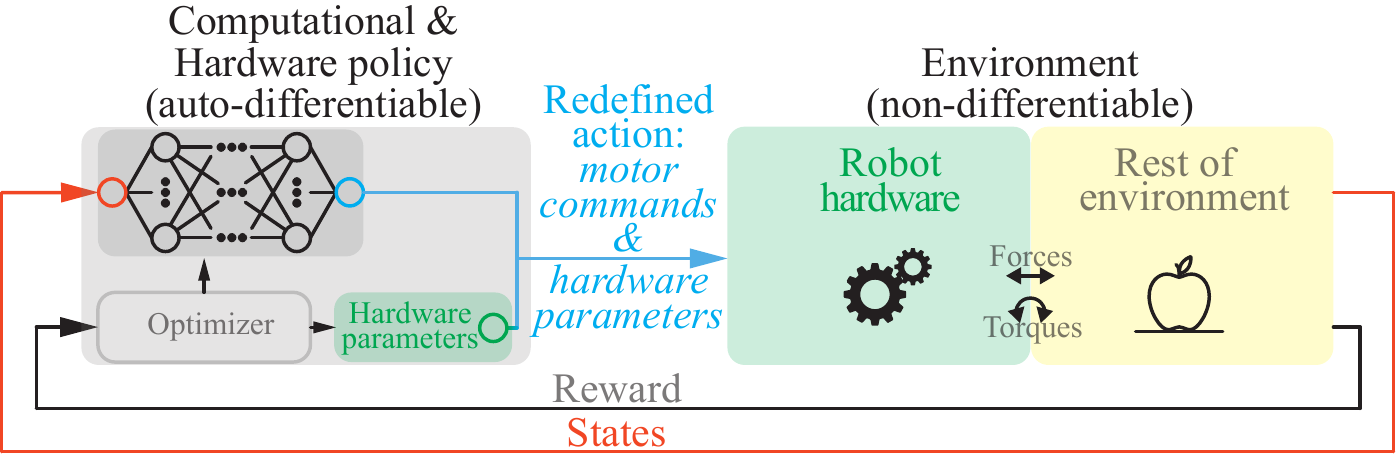}
        \caption{Hardware as Policy-Minimal}
        \label{fig:hw_as_action}
        \vspace{-6mm}
\end{figure*}

\textbf{Minimal Implementation (HWasP-Minimal).} In the general case, where should the split between the (differentiable) hardware policy and the (non-differentiable) rest of the environment simulation be performed? In particular, what if the hardware we care to optimize does not lend itself to a differentiable simulation using existing methods? 

Even in such a case, we argue that a ``minimal'' hardware policy is always possible: we can simply put the hardware parameters into the output layer of the original computational policy. In this case, $\bm{a}^{new}=[\bm{a},\bm{\phi}]^T$. Here, the policy gradient with respect to the hardware parameters is trivial but can be still useful to guide the update of parameters. When this case in implemented in practice, the transition function $\mathcal{F}(\bm{s}' | \bm{s}, \bm{a}^{new})$ typically operates in two steps: first, it sets the new values of the hardware parameters to the underlying simulator, then advances the simulation to the next step.

We illustrate HWasP-Minimal in Fig.~\ref{fig:hw_as_action}, which can be compared to HWasP as illustrated in Fig.~\ref{subfig:hw_as_policy}. HWasP-Minimal is simple to implement since it does not require a physics-based auto-differentiable hardware policy. As shown in our results, we found that HWasP-Minimal performs at least as well as or better than our baselines, but still below HWasP. 



\textbf{Comparison Baselines.} We compare HWasP and HWasP-Minimal with the following baselines:
\begin{compactitem}
\item CMA-ES/ARS with RL inner loop: we treat hardware parameters as hyperparameters, optimized in an outer loop using an evolutionary algorithm while the computational policy is learned (by RL algorithms such as PPO or TRPO) in an inner loop, for each set of hardware parameters. We tested two evolutionary algorithms: Covariance Matrix Adaptation - Evolution Strategy (CMA-ES) \cite{hansen2001completely}, and Augmented Random Search (ARS) \cite{Kumar_Tiwari_2019}. While ARS was originally tested on linear models, it can also be applied to non-linear systems (like our computational and mechanical policies) where it still represents a useful baseline.
\item CMA-ES/ARS: we use CMA-ES/ARS as gradient-free evolutionary strategies to directly learn both computational policy and hardware parameters, without a separate inner loop.
\end{compactitem}

\vspace{-2mm}
\section{A Mass-spring Toy Problem}
\label{sec:case_study_1}
\vspace{-2mm}

\begin{figure}[t!]
\centering

\begin{subfigure}[b]{0.39\textwidth}
\begin{tabular}{cc}
     \multicolumn{2}{c}{\includegraphics[width=0.5\linewidth]{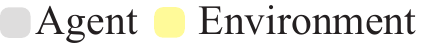}} \\
     \includegraphics[width=0.5\linewidth]{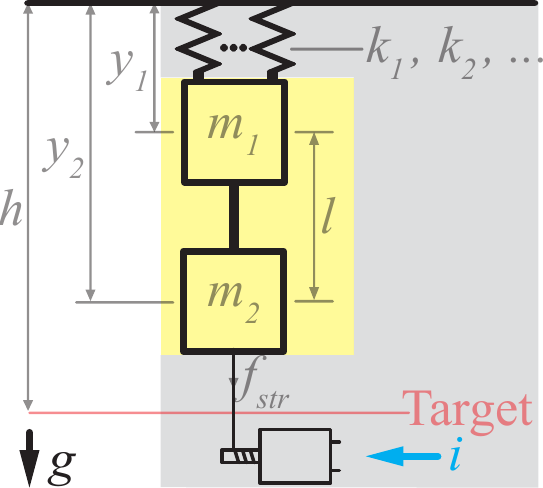}  & \hspace{-5mm} \includegraphics[width=0.5\linewidth]{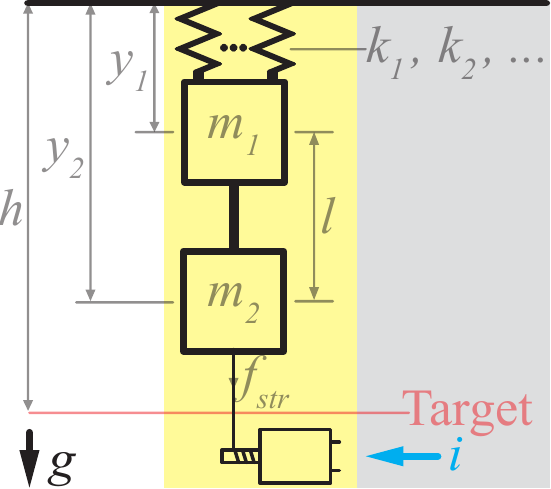} \\

     \vspace{-1mm}
\end{tabular}
\label{subfig:toy_problem_env}
\caption{Problem descriptions, left: HWasP, right: HWasP-Minimal.}
\vspace{-3mm}
\end{subfigure}
\begin{subfigure}[b]{0.60\textwidth}
\begin{tabular}{cc}
     \multicolumn{2}{c}{\includegraphics[width=0.8\linewidth]{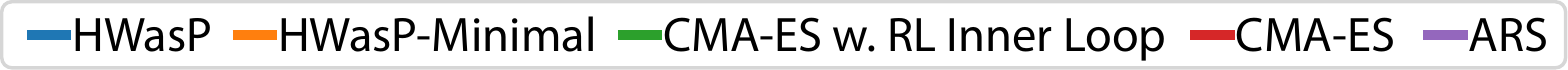}} \vspace{-1mm}\\
     \includegraphics[trim={0mm 0mm 0mm 10mm},clip,width=0.54\linewidth]{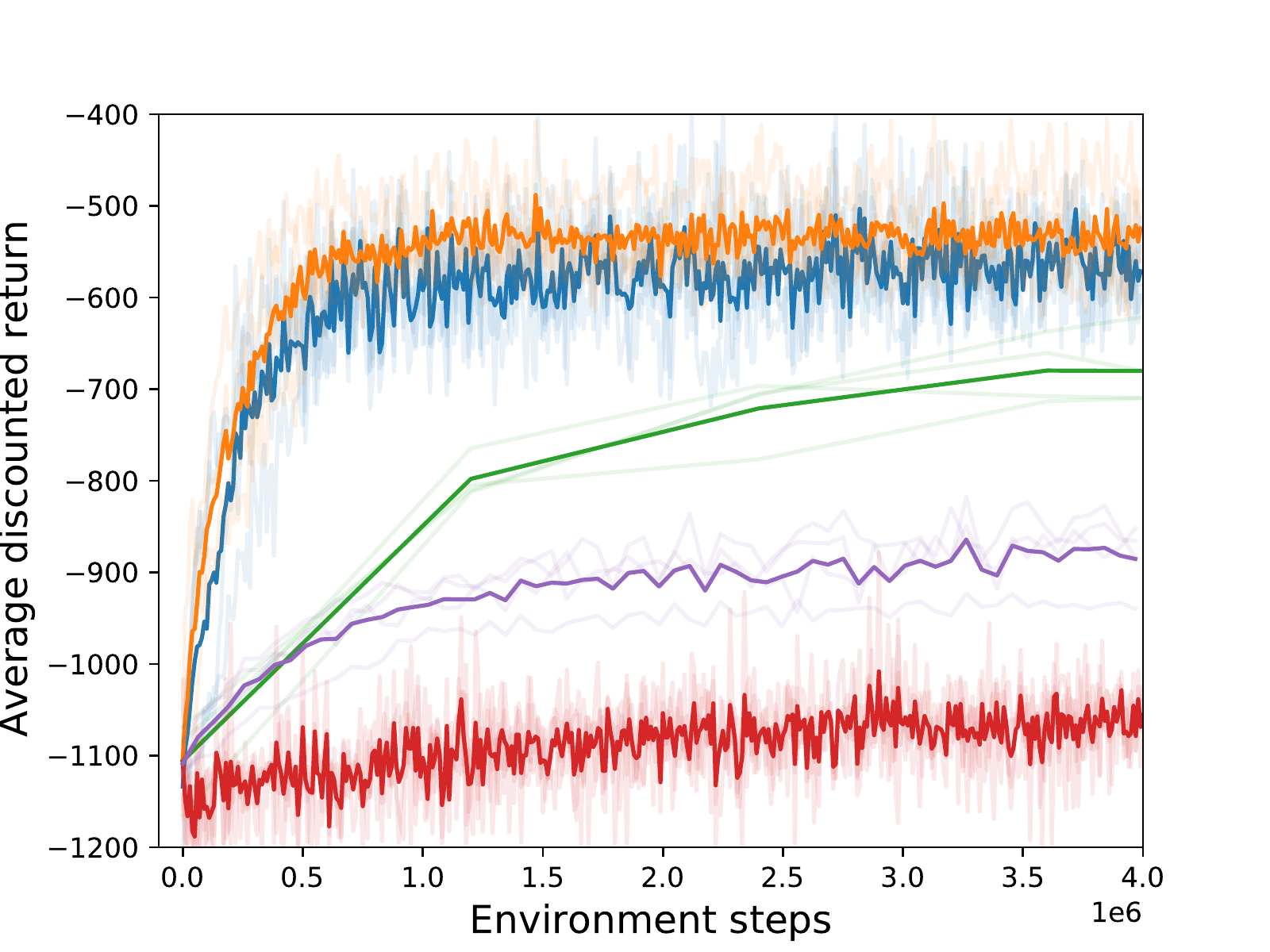}  &  \hspace{-9mm} \includegraphics[trim={5mm 0mm 0mm 10mm},clip,width=0.52\linewidth]{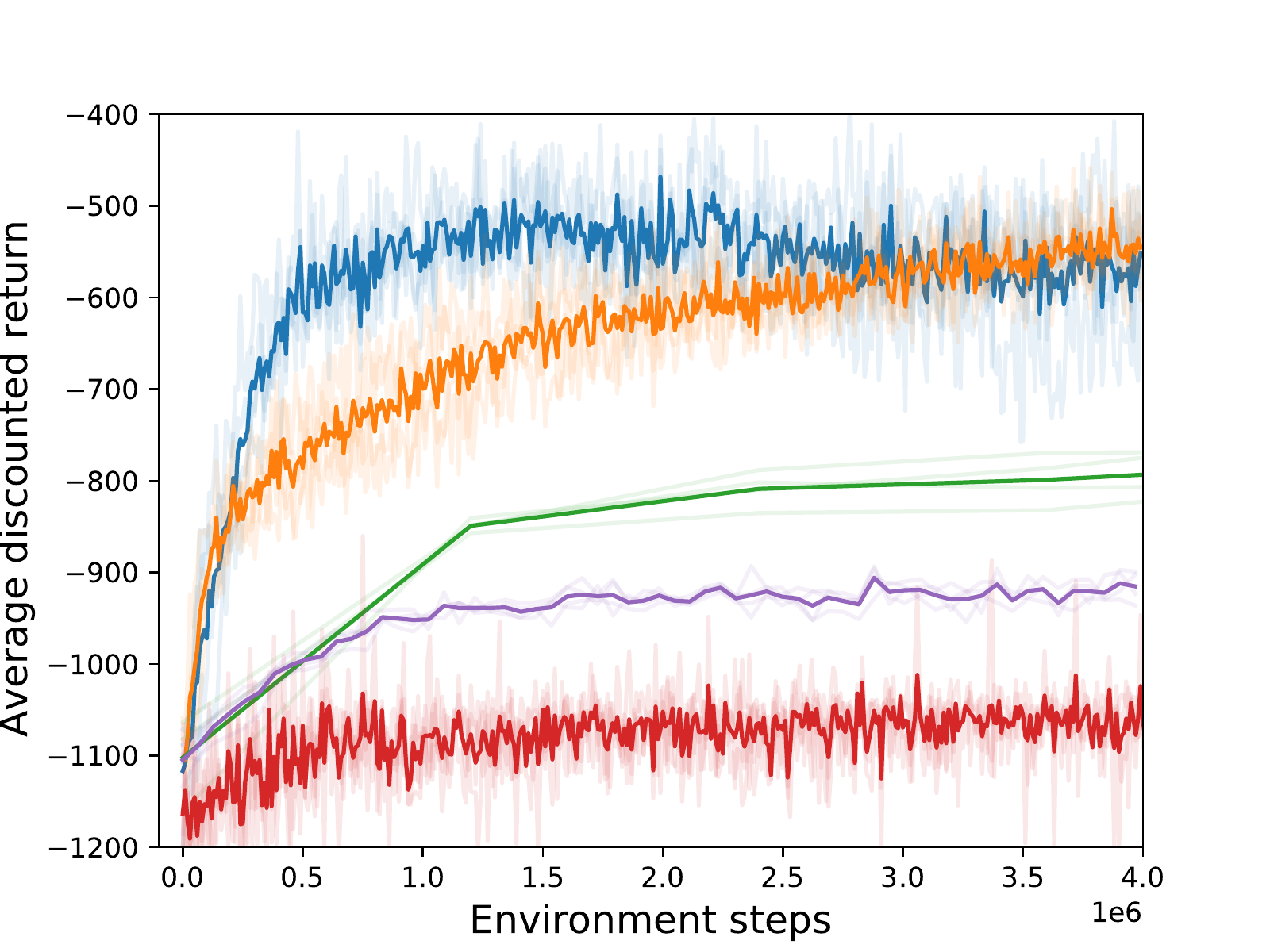}\\
     \vspace{-7mm}
\end{tabular}
\label{subfig:toy_problem_training_curves}
\caption{Training curves, left: 10 parameters, right: 50 parameters. \\ ~}
\vspace{-3mm}
\end{subfigure}
\caption{The mass-spring toy problem.}
\label{fig:toy_problems}
\vspace{-7mm}
\end{figure}

We present a one-dimensional example on the mass-spring system in Fig.~\ref{fig:toy_problems}a. Two point masses, connected by a massless bar, are hanging under $n$ parallel springs with stiffnesses $k_1, \ldots, k_n$. A motor can pull the lower mass by a string. The behavior is governed by a computational policy regulating the motor current, and by the hardware parameters (spring stiffnesses). We note that only the sum of spring stiffnesses matters since they are in parallel, but we still consider each stiffness as an individual parameter to test how our methods scale up for higher-dimensional problems. 

The input to the computational policy consists of $y_2$ and $\dot{y_2}$, and the output is motor current $i$. The goal is to optimize both the computational policy and the hardware parameters $k_1, \ldots, k_n$ such that the lower mass goes to the red target line ($y_2=h$) and stay there with minimum motor effort. (The exact formulation for the reward function we use is presented in Supplementary Materials \ref{appendix:toy_problem}.)

\textbf{Hardware as Policy.} In this case, we include the effect of the parallel springs in the mechanical policy. Using Hooke's Law, we model spring effects as a computational graph, with $\bm{k} = [k_1, \ldots, k_n]$ as parameters. The redefined action $\bm{a}^{new}$ consists of the total resultant force $f_{total} = f_{str} - f_{spr}$. The transition function $\mathcal{F}$ (rest of the environment) implements Newton's Law for the two masses, assuming $f_{total}$ as an external force. Details about the implementation of the computational graph can be found in Fig. \ref{fig:toy_problem_comp_graphs} in the Supplementary Materials.

\textbf{Hardware as Policy --- Minimal.} In this method, we simply re-define the action vector to also include spring stiffnesses: $\bm{a}^{new}=[i ,k_1, \cdots, k_n]^T$. The transition function $\mathcal{F}$ is responsible for modeling the dynamics of the springs and the two masses. 

\textbf{Results.} Fig.~\ref{fig:toy_problems}b shows the comparison of the training curves for both implementations of our method, as well as other baselines, for two cases: 10 and 50 parallel springs. In both cases, HWasP learns an effective joint policy that moves the lower mass to the target position. HWasP-Minimal works equally well for the smaller problem, but suffers a drop in performance as the number of hardware parameters increases. CMA-ES with RL inner loop also learns a joint policy to some extent, but learns slower than our method, especially for the larger problem. CMA-ES by itself does not exhibit any learning behavior over the number of samples tested. ARS achieves better performance than CMA-ES but is still worse than HWasP, HWasP-Minimal and CMA-ES with RL inner loop. For the numerical results of the optimized stiffnesses, please refer to the Supplementary Materials.

To test if our method can also optimize geometric parameters, we also performed co-optimization of the bar length and the control policy. This additional design case is detailed in Supplementary Materials.

\vspace{-2mm}
\section{Co-Design of an Underactuated Hand}
\label{sec:case_study_2}
\begin{figure}[t]
\begin{tabular}{cc}
\begin{tabular}{c}
\includegraphics[width=0.23\linewidth]{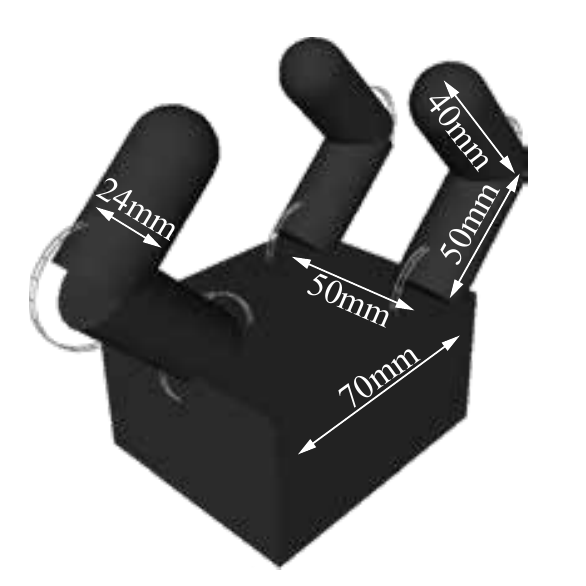} \vspace{-2mm}

\\
\includegraphics[width=0.23\linewidth]{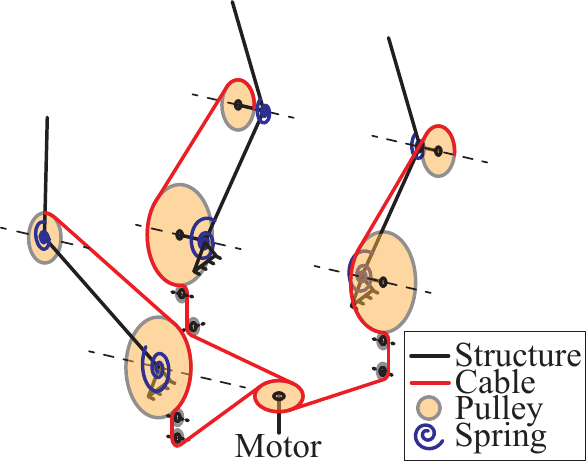}
\end{tabular}
&
\setlength{\tabcolsep}{0mm}
\begin{tabular}{cccc}
\includegraphics[width=0.18\linewidth]{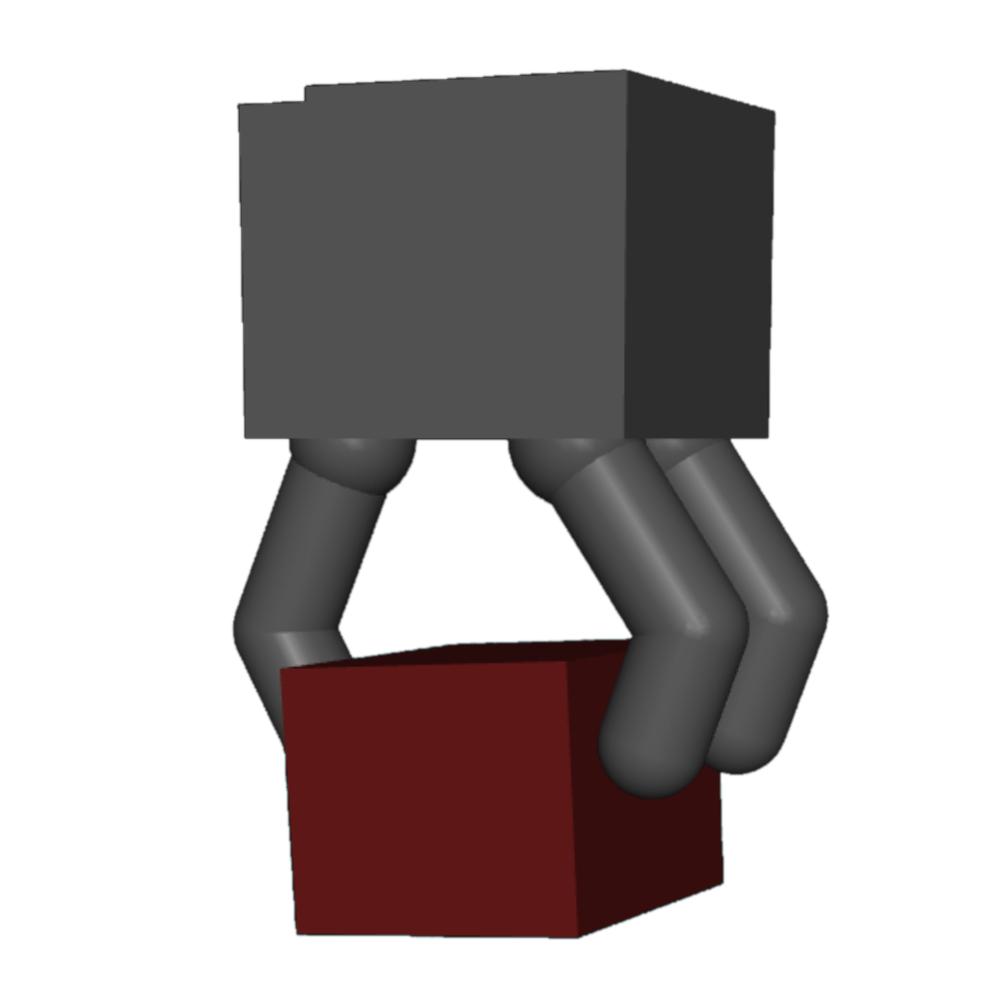}
&
\includegraphics[width=0.18\linewidth]{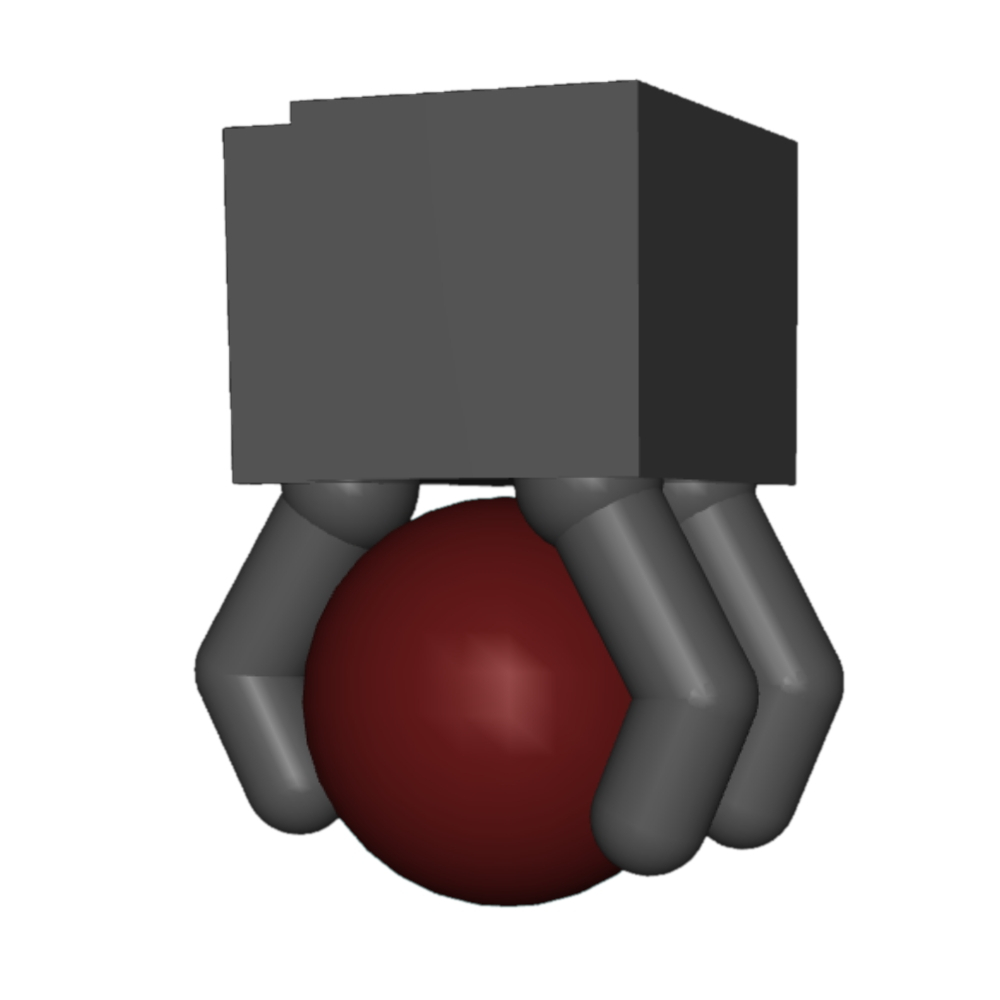}
&
\includegraphics[width=0.18\linewidth]{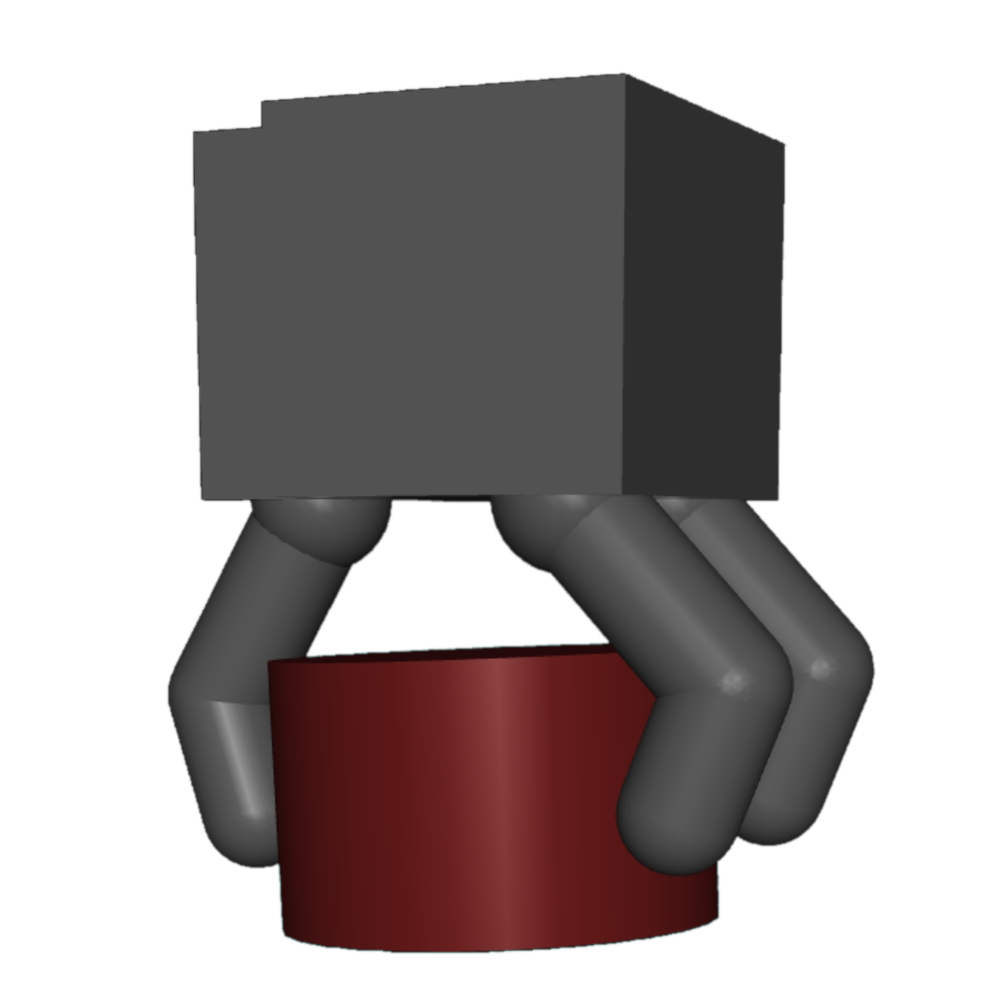}
&
\includegraphics[width=0.17\linewidth]{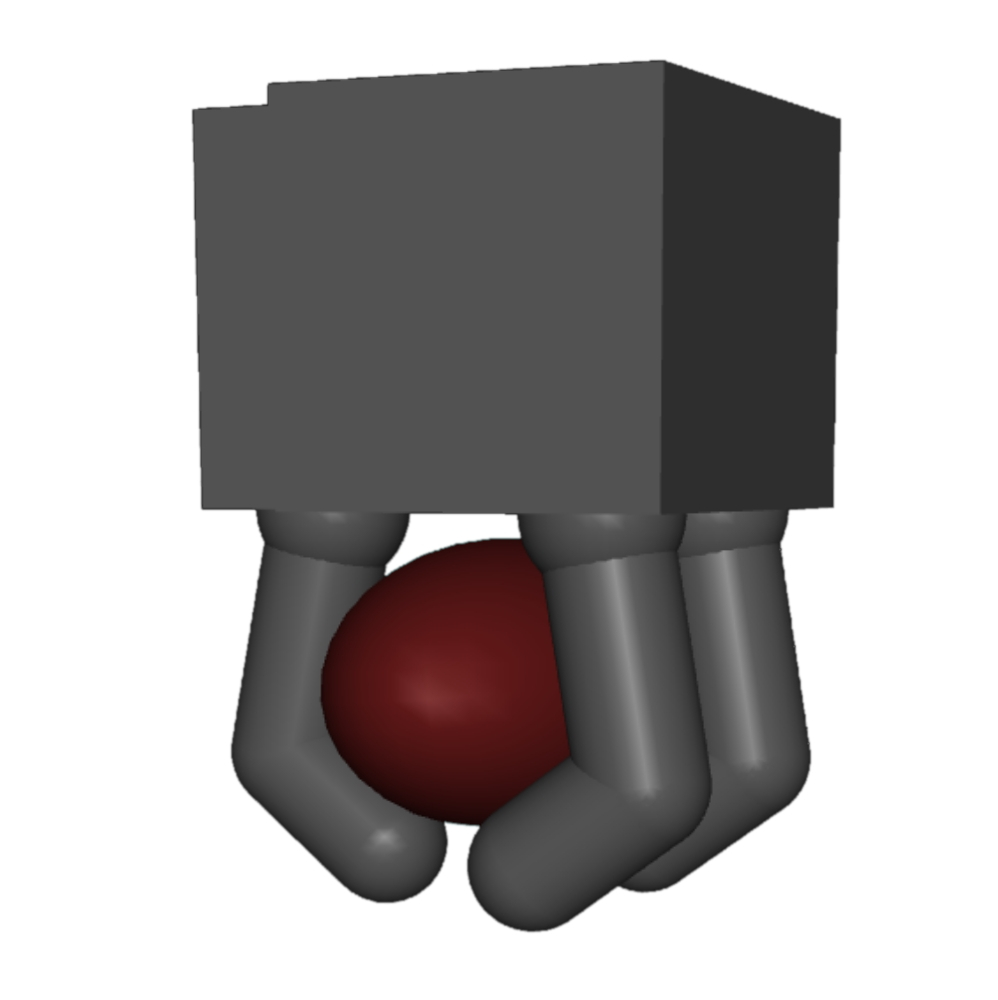}
\\
\includegraphics[width=0.18\linewidth]{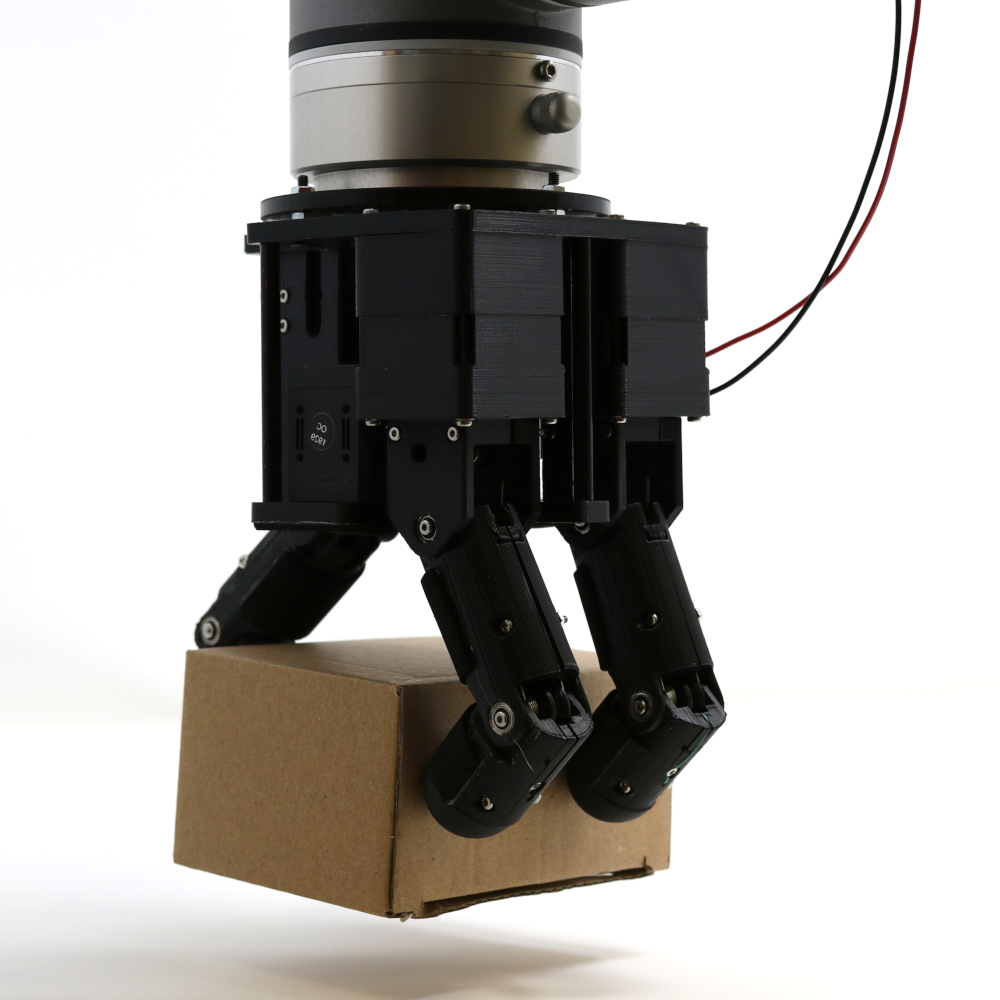}
&
\includegraphics[width=0.18\linewidth]{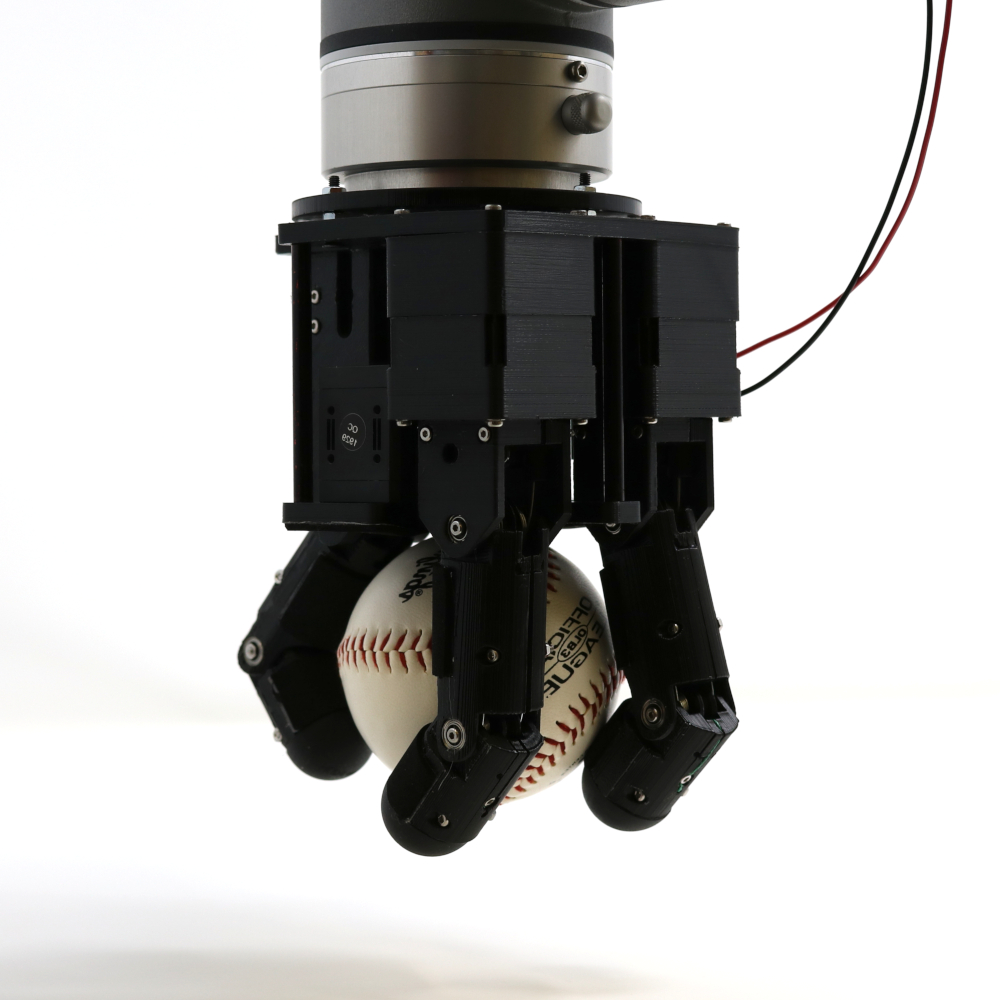}
&
\includegraphics[width=0.18\linewidth]{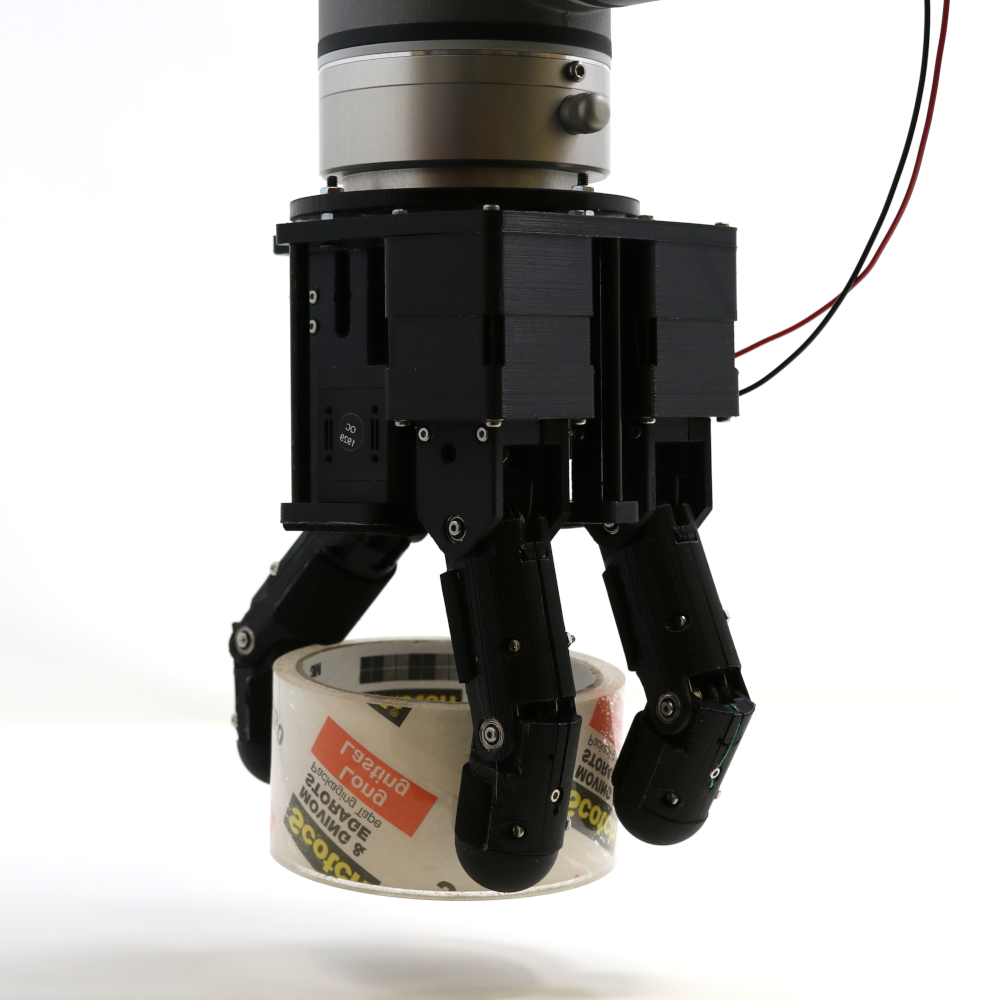}
&
\includegraphics[width=0.18\linewidth]{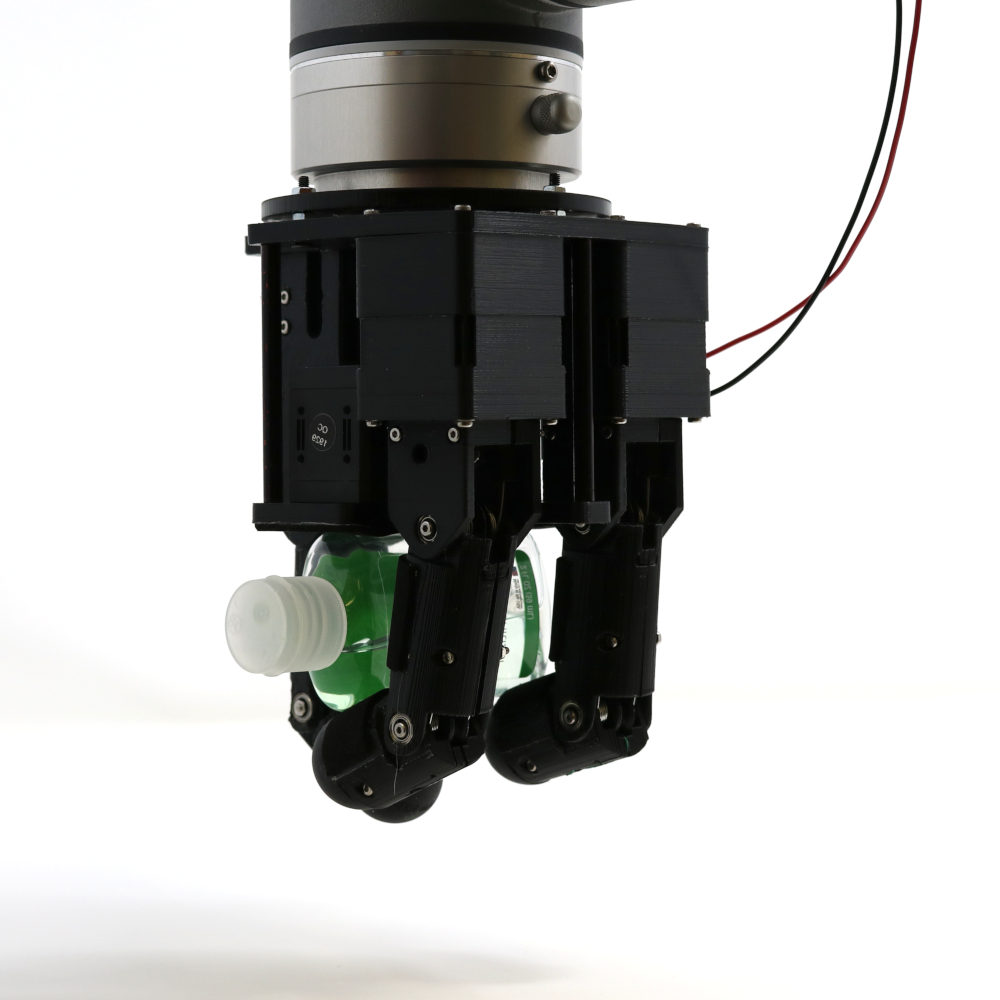}
\end{tabular}
\end{tabular}
\vspace{-2mm}
\caption{Hand design optimization problem. Left: hand kinematics, dimension, and tendon routing. Right: successful grasps executed in simulation and on a real hand prototype.}
\label{fig:nasa_hand_grasps}
\vspace{-7mm}
\end{figure}

\vspace{-2mm}
In this section we show how HWasP can be applied to a real-world design problem: optimizing the mechanism and the control policy for an underactuated robot hand. The high-level design goal, inspired by previous work~\cite{chen2020underactuation}, is to design a robot hand that is compact, but still versatile (able to grasp different shaped objects, as shown in Fig~\ref{fig:nasa_hand_grasps}). To achieve the stated compactness goal, all joints are driven by a single motor, via an underactuated transmission mechanism: one motor actuates all joints by tendons in the flexion direction (see Fig~\ref{fig:nasa_hand_grasps}). Finger extension is passive, via preloaded torsional springs. The mechanical parameters that govern the behavior of this mechanism consist of tendon pulley radii in each joint, as well as stiffness values and preload angles for restoring springs. 

Here, we look to simultaneously optimize the hardware parameters along with a computational policy that determines how to position the hand and use the hand motor. The input to the computational policy consists of palm and object positions, object size, and current motor travel and torque. Its output contains a relative position setpoint for hand motor travel as well as palm position commands.

From a hardware perspective, we aim to optimize all the underactuated transmission parameters listed above. We note that, in this work, we do not try to optimize the kinematic structure or topology for the hand. Unlike the underactuated transmission, these aspects do not lend themselves to parameterization and implementation as computational graphs, preventing the use of the HWasP method in its current form. While HWasP-Minimal could still be applied, we leave that for future investigations.

We tested our method with two grasping tasks: top-down grasping with only z-axis motion for the palm movement (Z-Grasp), and top-down grasping with 3-dimensional palm motion (3D-Grasp). The former is a simplified problem version of the latter, and easier to train. Since hardware parameters can be large in scale comparing to weights and biases in the neural network, a small change can lead to a large shift of the joint policy output distribution during training, which in turn can result in local optimum in the reward landscape. To alleviate this issue, we add scaling factors for parameters in the hardware policy computational graph. We use TRPO \cite{schulman2015trust} as the underlying RL algorithm as it allows for hard constraints on action distribution changes. Additional details on problem formulation and training can be found in Supplementary Materials.

We also apply Domain Randomization \cite{tobin2017domain} in the training to increase the chance of successful sim-to-real transfer. We randomized object shape, size, weight, friction coefficient and inertia, injected sensor and actuation noise, and applied random disturbance wrenches on the hand-object system.

\textbf{Hardware as Policy.} In this case, we model the complete underactuated transmission as a computational graph and include it in our mechanical policy. The input to the mechanical policy consists of the commanded motor travel (output by the computational policy), as well current joint angles. Its output consists of hand joint torques. To perform this computation, we use a tendon model that computes the elongation of the tendon in response to motor travel and joint positions, then uses that value to compute tendon forces and joint torques. Details of this model as well as its implementation as an auto-differentiable computational graph can be found in Supplementary Materials \ref{appendix:hand}.

The redefined action $\bm{a}^{new}$ contains the palm position command output by the computational policy, and the joint torques produced by the mechanical policy. The rest of the environment comprises the hand-object system without the tendon underactuation mechanisms, i.e. with independent joints.

\textbf{Hardware as Policy --- Minimal}. In this case, all hardware parameters are simply appended to the output of the computational policy. The underactuated transmission model is part of the environment, along with the rest of the hand as well as the object.

\begin{figure}[t!]
    \hspace{-2mm}
    \begin{subfigure}[b]{0.55\linewidth}
        \includegraphics[width=\linewidth]{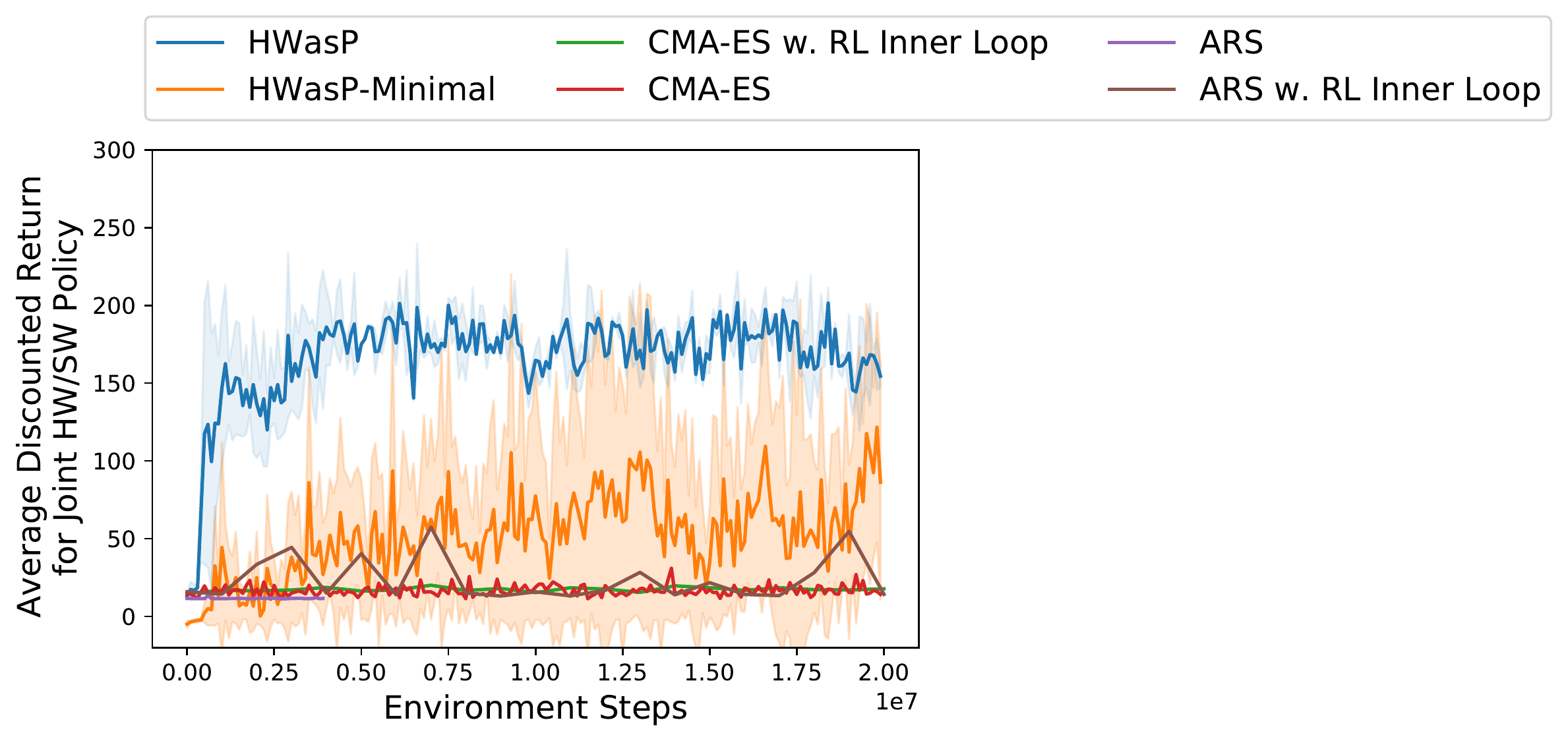}
    \end{subfigure}
    \hspace{-32mm}
    \begin{subfigure}[b]{0.30\linewidth}
        \includegraphics[width=\linewidth]{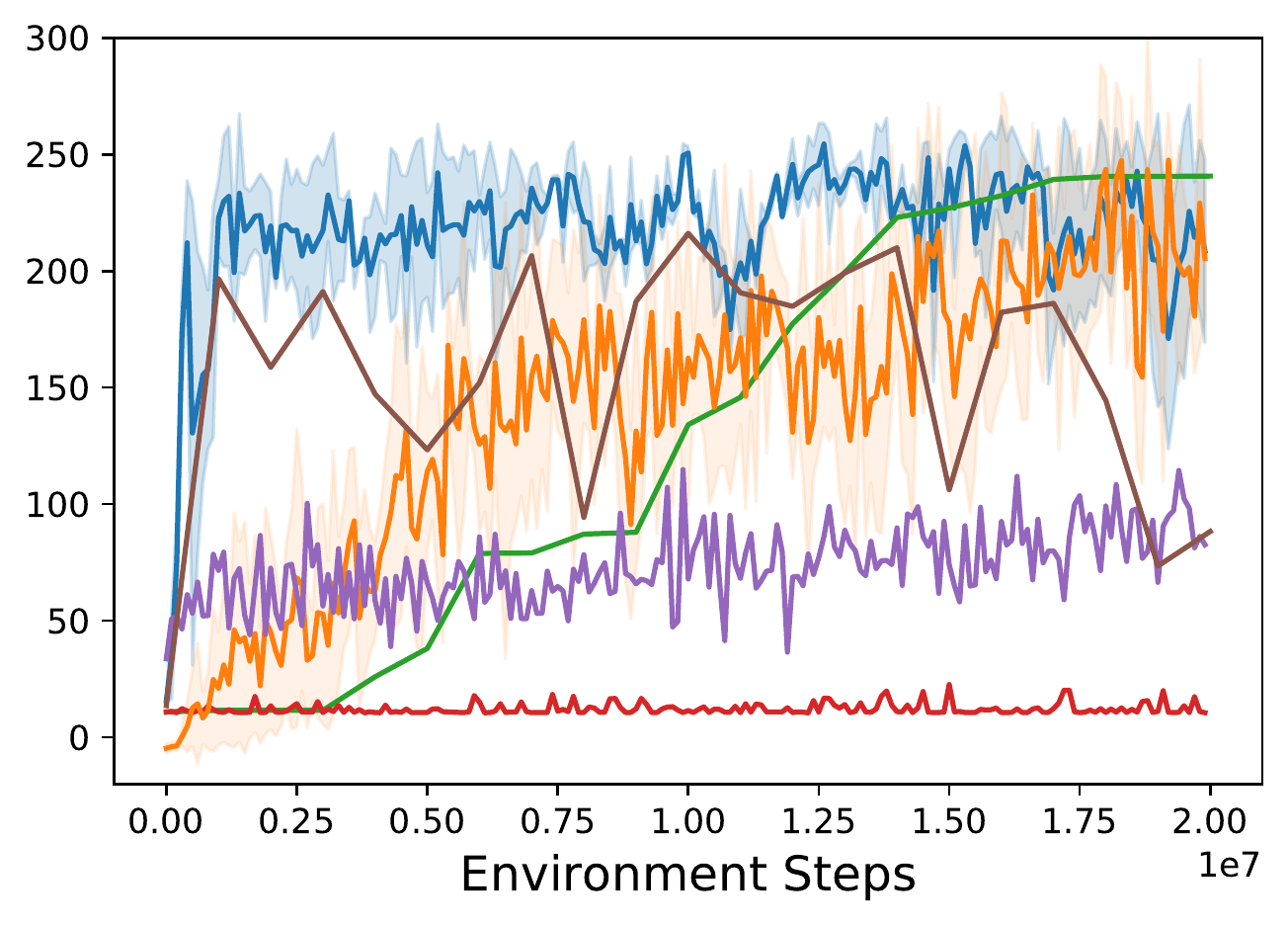}
    \end{subfigure}
    \hspace{-2mm}
    \begin{subfigure}[b]{0.30\linewidth}
        \includegraphics[width=\linewidth]{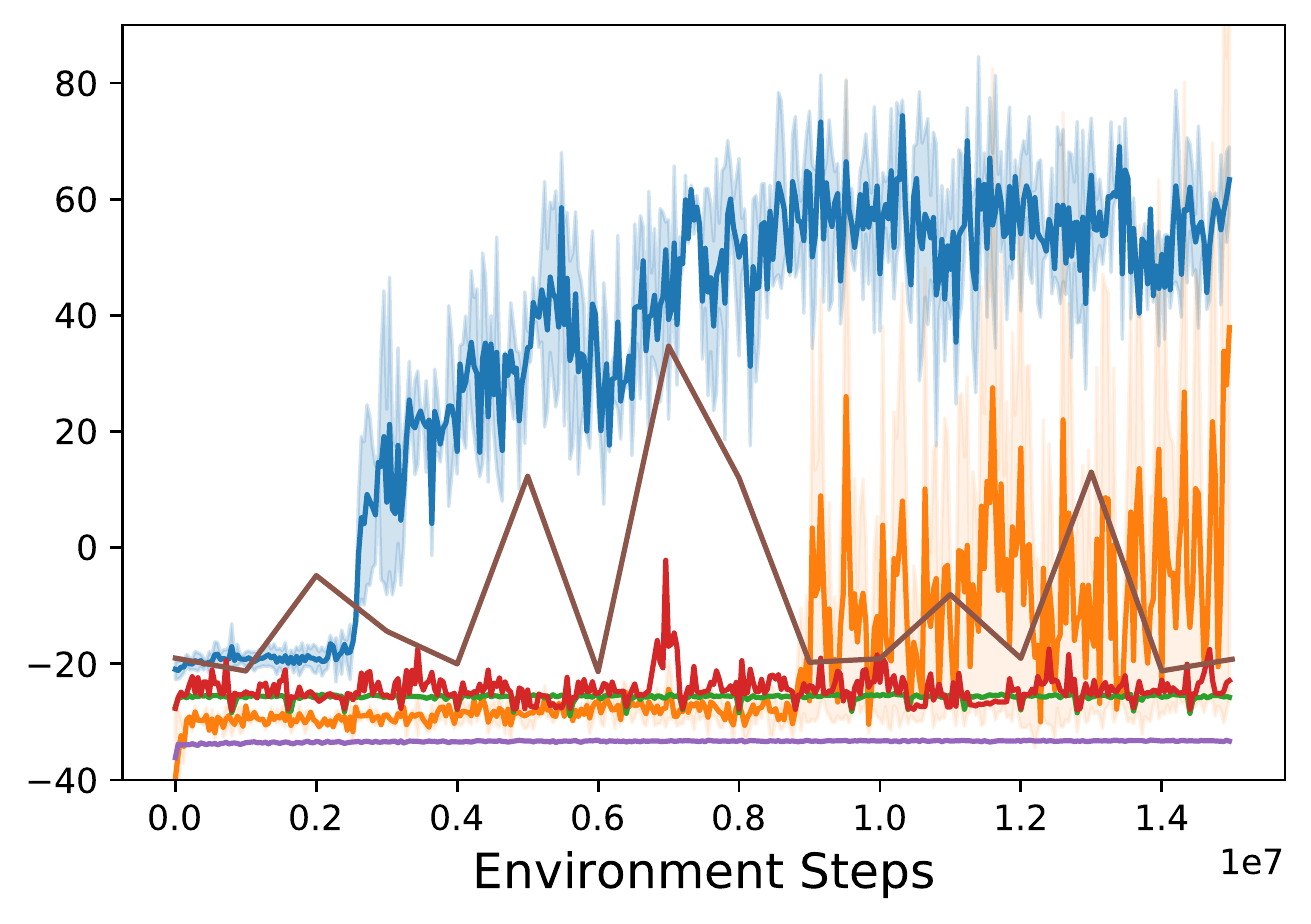}
    \end{subfigure}
    \caption{Training curves for the grasping problem. Left: Z-Grasp with a large hardware parameter search range. Middle: Z-Grasp with a small hardware search range. Right: 3D-Grasp with a small search range. }
    \label{fig:nasa_hand_training_curves}
    \vspace{-6mm}
\end{figure}

\textbf{Results.} Our results are shown in Fig. \ref{fig:nasa_hand_training_curves}. For Z-Grasp (left plot), HWasP learns an effective computational/hardware policy. HWasP-Minimal does the same, but at a slower pace. Neither evolutionary strategy shows any learning behavior over a similar number of training steps.

To gain additional insight, we also tried an easier version of the same problem with the search range for the hardware parameters reduced by a factor of 8 (middle plot). Here, some of our baselines can also learn effective policies, particularly with an RL inner loop, but HWasP is still the most efficient. 

Finally, we investigated performance for the more complex 3D-Grasp task. With a large search range, neither method was able to learn. However, with a reduced search range, HWasP learned an effective policy, while neither CMA-ES-based method displayed any learning behavior over a similar timescale. The values of the optimized hardware parameters are shown in the Supplementary Materials. 

\textbf{Validation with Physical Prototype.} To validate our results in the real world, we physically built the hand with the parameters resulted from the co-optimization. The hand is 3D printed, and actuated by a single position-controlled servo motor. Fig. \ref{fig:nasa_hand_grasps} shows grasps obtained by this physical prototype, compared to their simulated counterparts. All shown grasps are stable and allow object lift and transport. We note that, as expected, the hand is highly versatile and can perform a wide range of both fingertip and enveloping grasps, for objects of varying shape and size. This shows that the optimized hardware policy is indeed effective in the real world.

Furthermore, we tested the combined hardware and computational policies for both Z-Grasp and 3D-Grasp on the real hand. Here, the computational policy determined how to position the hand (implemented in practice using a UR5 robot with position control) and also issued all commands to the hand servo. While driving the optimized hand, the computational policy achieved $100\%$ success rate for Z-Grasp with boxes and balls and $90\%$ with cylinders. For 3D-Grasp, the success rate was $100\%$ on boxes, $80\%$ on cylinders and $50\%$ on balls. We note however that the computational policy expects object pose as input, which was not available on our experimental setup; instead, we started each run with the object in a known location, which was provided as part of the observation. However, on the ball object, early contacts occasionally caused untracked object displacement, which lowered the success rate. While we hope to train and test more complex computational policies in the future, including with real sensor feedback, we believe these experiments underlined the effectiveness of the hardware policy, and its ability to operate together with the jointly optimized computational policy.


\vspace{-2mm}
\section{Disscussion and Conclusion}
\vspace{-2mm}
\label{sec:discussion_and_conclusion}

Our results show that the HWasP approach is able to learn combined computational and mechanical policies. We attribute this performance to the fact that HWasP connects different hardware parameters via a computational graph based on the laws of physics, and can provide the physics-based gradient of the action probability w.r.t the hardware parameters. The HWasP-Minimal implementation does not provide such information, and the policy gradient can only be estimated via sampling, which is usually less efficient, particularly for high-dimensional problems. In consequence, HWasP-Minimal also shows the ability to learn effective policies, but with reduced performance.

Compared to gradient-free evolutionary baselines for joint hardware-software co-optimization, HWasP always learns faster, while HWasP-Minimal is at least as effective as the best baseline algorithm. We note that combining an RL inner loop for the computational policy with a CMA-ES outer loop for hardware parameters proved more effective than directly using CMA-ES for the complete problem. Still, HWasP outperforms both methods. 

The biggest advantage of HWasP-Minimal is that, like gradient-free methods, it does not depend on auto-differentiable physics, and is widely applicable with straightforward implementations to various problems using existing non-differentiable physics engines. We believe that our methods represent a step towards a framework where an algorithm designer can "tune the slider" to decide how much physics to include in the computational policy, based on the trade-offs between computation efficiency, ease of development, and the availability of auto-differentiable physics simulations.

In its current stage, our work still presents a number of limitations. The computational aspects of the policies we have explored so far are relatively simple (e.g. limiting hand motion to 1- or 3-DOF). We hope to explore more challenging robotic tasks, including 6-DOF hand positioning. Here, we also used HWasP with a single task, but believe it is possible to learn more robust hardware parameters by trying to generalize to multiple tasks.
Finally, we aim to include additional hardware aspects in the optimization, such as mechanism kinematics, morphology, or link dimensions.

We believe the proposed idea of considering hardware as part of the policy can enable co-design of hardware and software using existing RL toolkits, with new computational graph structures but little change in the learning algorithms. We hope this work can open new opportunities for task-based hardware-software co-design of intelligent systems, for researchers in both RL and hardware design.


\acknowledgments{The authors would like to thank Roberto Calandra for insightful discussions and suggestions. This work was supported in part by ONR Grant N00014-16-1-2026 and by a Google Faculty Award.}

\clearpage

\bibliography{bib/analysis,bib/biomechanics,bib/control,bib/design,bib/learning,bib/planning,bib/sensing,bib/simulation}

\begin{thebibliography}{36}
\providecommand{\natexlab}[1]{#1}
\providecommand{\url}[1]{\texttt{#1}}
\expandafter\ifx\csname urlstyle\endcsname\relax
  \providecommand{\doi}[1]{doi: #1}\else
  \providecommand{\doi}{doi: \begingroup \urlstyle{rm}\Url}\fi

\bibitem[Lichtwark and Wilson(2007)]{lichtwark2007achilles}
G.~Lichtwark and A.~Wilson.
\newblock Is achilles tendon compliance optimised for maximum muscle efficiency
  during locomotion?
\newblock \emph{Journal of biomechanics}, 40\penalty0 (8):\penalty0 1768--1775,
  2007.

\bibitem[Santello et~al.(2013)Santello, Baud-Bovy, and
  J{\"o}rntell]{santello2013neural}
M.~Santello, G.~Baud-Bovy, and H.~J{\"o}rntell.
\newblock Neural bases of hand synergies.
\newblock \emph{Frontiers in computational neuroscience}, 7:\penalty0 23, 2013.

\bibitem[Hammami et~al.(2016)Hammami, Chaouachi, Makhlouf, Granacher, and
  Behm]{hammami2016associations}
R.~Hammami, A.~Chaouachi, I.~Makhlouf, U.~Granacher, and D.~G. Behm.
\newblock Associations between balance and muscle strength, power performance
  in male youth athletes of different maturity status.
\newblock \emph{Pediatric Exercise Science}, 28\penalty0 (4):\penalty0
  521--534, 2016.

\bibitem[Young(2003)]{young2003evolution}
R.~W. Young.
\newblock Evolution of the human hand: the role of throwing and clubbing.
\newblock \emph{Journal of Anatomy}, 202\penalty0 (1):\penalty0 165--174, 2003.

\bibitem[Rajeswaran et~al.(2017)Rajeswaran, Kumar, Gupta, Vezzani, Schulman,
  Todorov, and Levine]{rajeswaran2017learning}
A.~Rajeswaran, V.~Kumar, A.~Gupta, G.~Vezzani, J.~Schulman, E.~Todorov, and
  S.~Levine.
\newblock Learning complex dexterous manipulation with deep reinforcement
  learning and demonstrations.
\newblock \emph{arXiv preprint arXiv:1709.10087}, 2017.

\bibitem[Andrychowicz et~al.(2018)Andrychowicz, Baker, Chociej, Jozefowicz,
  McGrew, Pachocki, Petron, Plappert, Powell, Ray,
  et~al.]{andrychowicz2018learning}
M.~Andrychowicz, B.~Baker, M.~Chociej, R.~Jozefowicz, B.~McGrew, J.~Pachocki,
  A.~Petron, M.~Plappert, G.~Powell, A.~Ray, et~al.
\newblock Learning dexterous in-hand manipulation.
\newblock \emph{arXiv preprint arXiv:1808.00177}, 2018.

\bibitem[Haarnoja et~al.(2018)Haarnoja, Ha, Zhou, Tan, Tucker, and
  Levine]{haarnoja2018learning}
T.~Haarnoja, S.~Ha, A.~Zhou, J.~Tan, G.~Tucker, and S.~Levine.
\newblock Learning to walk via deep reinforcement learning.
\newblock \emph{arXiv preprint arXiv:1812.11103}, 2018.

\bibitem[Birglen and Gosselin(2004)]{birglen2004kinetostatic}
L.~Birglen and C.~M. Gosselin.
\newblock Kinetostatic analysis of underactuated fingers.
\newblock \emph{IEEE Transactions on Robotics and Automation}, 20\penalty0
  (2):\penalty0 211--221, 2004.

\bibitem[Odhner et~al.(2014)Odhner, Jentoft, Claffee, Corson, Tenzer, Ma,
  Buehler, Kohout, Howe, and Dollar]{odhner2014compliant}
L.~U. Odhner, L.~P. Jentoft, M.~R. Claffee, N.~Corson, Y.~Tenzer, R.~R. Ma,
  M.~Buehler, R.~Kohout, R.~D. Howe, and A.~M. Dollar.
\newblock A compliant, underactuated hand for robust manipulation.
\newblock \emph{The International Journal of Robotics Research}, 33\penalty0
  (5):\penalty0 736--752, 2014.

\bibitem[Tobin et~al.(2017)Tobin, Fong, Ray, Schneider, Zaremba, and
  Abbeel]{tobin2017domain}
J.~Tobin, R.~Fong, A.~Ray, J.~Schneider, W.~Zaremba, and P.~Abbeel.
\newblock Domain randomization for transferring deep neural networks from
  simulation to the real world.
\newblock In \emph{2017 IEEE/RSJ International Conference on Intelligent Robots
  and Systems (IROS)}, pages 23--30. IEEE, 2017.

\bibitem[Schulman et~al.(2015)Schulman, Levine, Abbeel, Jordan, and
  Moritz]{schulman2015trust}
J.~Schulman, S.~Levine, P.~Abbeel, M.~Jordan, and P.~Moritz.
\newblock Trust region policy optimization.
\newblock In \emph{Intl. Conf. on machine learning}, pages 1889--1897, 2015.

\bibitem[Schulman et~al.(2017)Schulman, Wolski, Dhariwal, Radford, and
  Klimov]{schulman2017proximal}
J.~Schulman, F.~Wolski, P.~Dhariwal, A.~Radford, and O.~Klimov.
\newblock Proximal policy optimization algorithms.
\newblock \emph{arXiv preprint arXiv:1707.06347}, 2017.

\bibitem[Lillicrap et~al.(2015)Lillicrap, Hunt, Pritzel, Heess, Erez, Tassa,
  Silver, and Wierstra]{lillicrap2015continuous}
T.~P. Lillicrap, J.~J. Hunt, A.~Pritzel, N.~Heess, T.~Erez, Y.~Tassa,
  D.~Silver, and D.~Wierstra.
\newblock Continuous control with deep reinforcement learning.
\newblock \emph{arXiv preprint arXiv:1509.02971}, 2015.

\bibitem[Park and Asada(1994)]{park1994concurrent}
J.-H. Park and H.~Asada.
\newblock Concurrent design optimization of mechanical structure and control
  for high speed robots.
\newblock \emph{Journal of dynamic systems, measurement, and control},
  116\penalty0 (3):\penalty0 344--356, 1994.

\bibitem[Paul and Bongard()]{paul2001road}
C.~Paul and J.~C. Bongard.
\newblock The road less travelled: Morphology in the optimization of biped
  robot locomotion.
\newblock In \emph{Proceedings 2001 IEEE/RSJ International Conference on
  Intelligent Robots and Systems. Expanding the Societal Role of Robotics in
  the the Next Millennium (Cat. No. 01CH37180)}, volume~1, pages 226--232.
  IEEE.

\bibitem[Geijtenbeek et~al.(2013)Geijtenbeek, Van De~Panne, and Van
  Der~Stappen]{geijtenbeek2013flexible}
T.~Geijtenbeek, M.~Van De~Panne, and A.~F. Van Der~Stappen.
\newblock Flexible muscle-based locomotion for bipedal creatures.
\newblock \emph{ACM Transactions on Graphics (TOG)}, 32\penalty0 (6):\penalty0
  206, 2013.

\bibitem[Ha et~al.(2018)Ha, Coros, Alspach, Kim, and
  Yamane]{ha2018computational}
S.~Ha, S.~Coros, A.~Alspach, J.~Kim, and K.~Yamane.
\newblock Computational co-optimization of design parameters and motion
  trajectories for robotic systems.
\newblock \emph{The International Journal of Robotics Research}, 37\penalty0
  (13-14):\penalty0 1521--1536, 2018.

\bibitem[Liao et~al.(2019)Liao, Wang, Yang, Lee, Pister, Levine, and
  Calandra]{liao2019data}
T.~Liao, G.~Wang, B.~Yang, R.~Lee, K.~Pister, S.~Levine, and R.~Calandra.
\newblock Data-efficient learning of morphology and controller for a
  microrobot.
\newblock In \emph{2019 International Conference on Robotics and Automation
  (ICRA)}, pages 2488--2494. IEEE, 2019.

\bibitem[Sims(1994)]{sims1994evolving}
K.~Sims.
\newblock Evolving virtual creatures.
\newblock In \emph{Proceedings of the 21st annual conference on Computer
  graphics and interactive techniques}, pages 15--22. ACM, 1994.

\bibitem[Lipson and Pollack(2000)]{lipson2000automatic}
H.~Lipson and J.~B. Pollack.
\newblock Automatic design and manufacture of robotic lifeforms.
\newblock \emph{Nature}, 406\penalty0 (6799):\penalty0 974, 2000.

\bibitem[Cheney et~al.(2014)Cheney, MacCurdy, Clune, and
  Lipson]{cheney2014unshackling}
N.~Cheney, R.~MacCurdy, J.~Clune, and H.~Lipson.
\newblock Unshackling evolution: evolving soft robots with multiple materials
  and a powerful generative encoding.
\newblock \emph{ACM SIGEVOlution}, 7\penalty0 (1):\penalty0 11--23, 2014.

\bibitem[Cheney and Lipson(2016)]{cheney2016topological}
N.~Cheney and H.~Lipson.
\newblock Topological evolution for embodied cellular automata.
\newblock \emph{Theoretical Computer Science}, 633:\penalty0 19--27, 2016.

\bibitem[Nygaard et~al.(2018)Nygaard, Martin, Samuelsen, Torresen, and
  Glette]{nygaard2018real}
T.~F. Nygaard, C.~P. Martin, E.~Samuelsen, J.~Torresen, and K.~Glette.
\newblock Real-world evolution adapts robot morphology and control to hardware
  limitations.
\newblock In \emph{Proceedings of the Genetic and Evolutionary Computation
  Conference}, pages 125--132, 2018.

\bibitem[Ha(2018)]{ha2018reinforcement}
D.~Ha.
\newblock Reinforcement learning for improving agent design.
\newblock \emph{arXiv preprint arXiv:1810.03779}, 2018.

\bibitem[Schaff et~al.(2019)Schaff, Yunis, Chakrabarti, and
  Walter]{schaff2019jointly}
C.~Schaff, D.~Yunis, A.~Chakrabarti, and M.~R. Walter.
\newblock Jointly learning to construct and control agents using deep
  reinforcement learning.
\newblock In \emph{IEEE Intl. Conf. on Robotics and Automation}, pages
  9798--9805. IEEE, 2019.

\bibitem[Vermeer et~al.(2018)Vermeer, Kuppens, and
  Herder]{vermeer2018kinematic}
K.~Vermeer, R.~Kuppens, and J.~Herder.
\newblock Kinematic synthesis using reinforcement learning.
\newblock In \emph{International Design Engineering Technical Conferences and
  Computers and Information in Engineering Conference}, volume 51753, page
  V02AT03A009. ASME, 2018.

\bibitem[Luck et~al.(2019)Luck, Ben~Amor, and Calandra]{luck2019coadapt}
K.~S. Luck, H.~Ben~Amor, and R.~Calandra.
\newblock Data-efficient co-adaptation of morphology and behaviour with deep
  reinforcement learning.
\newblock In \emph{Conf. on Robot Learning}, 2019.

\bibitem[de~Avila Belbute-Peres et~al.(2018)de~Avila Belbute-Peres, Smith,
  Allen, Tenenbaum, and Kolter]{de2018end}
F.~de~Avila Belbute-Peres, K.~Smith, K.~Allen, J.~Tenenbaum, and J.~Z. Kolter.
\newblock End-to-end differentiable physics for learning and control.
\newblock In \emph{Advances in Neural Information Processing Systems}, pages
  7178--7189, 2018.

\bibitem[Degrave et~al.(2019)Degrave, Hermans, Dambre, and
  Wyffels]{degrave2019differentiable}
J.~Degrave, M.~Hermans, J.~Dambre, and F.~Wyffels.
\newblock A differentiable physics engine for deep learning in robotics.
\newblock \emph{Frontiers in neurorobotics}, 13:\penalty0 6, 2019.

\bibitem[Hu et~al.(2019{\natexlab{a}})Hu, Liu, Spielberg, Tenenbaum, Freeman,
  Wu, Rus, and Matusik]{hu2019chainqueen}
Y.~Hu, J.~Liu, A.~Spielberg, J.~B. Tenenbaum, W.~T. Freeman, J.~Wu, D.~Rus, and
  W.~Matusik.
\newblock Chainqueen: A real-time differentiable physical simulator for soft
  robotics.
\newblock In \emph{2019 International Conference on Robotics and Automation
  (ICRA)}, pages 6265--6271. IEEE, 2019{\natexlab{a}}.

\bibitem[Hu et~al.(2019{\natexlab{b}})Hu, Anderson, Li, Sun, Carr,
  Ragan-Kelley, and Durand]{hu2019difftaichi}
Y.~Hu, L.~Anderson, T.-M. Li, Q.~Sun, N.~Carr, J.~Ragan-Kelley, and F.~Durand.
\newblock Difftaichi: Differentiable programming for physical simulation.
\newblock \emph{arXiv preprint arXiv:1910.00935}, 2019{\natexlab{b}}.

\bibitem[Todorov et~al.(2012)Todorov, Erez, and Tassa]{todorov2012mujoco}
E.~Todorov, T.~Erez, and Y.~Tassa.
\newblock Mujoco: A physics engine for model-based control.
\newblock In \emph{2012 IEEE/RSJ International Conference on Intelligent Robots
  and Systems}, pages 5026--5033. IEEE, 2012.

\bibitem[Hansen and Ostermeier(2001)]{hansen2001completely}
N.~Hansen and A.~Ostermeier.
\newblock Completely derandomized self-adaptation in evolution strategies.
\newblock \emph{Evolutionary computation}, 9\penalty0 (2):\penalty0 159--195,
  2001.

\bibitem[Kumar~Tiwari and Nadimpalli(2019)]{Kumar_Tiwari_2019}
A.~Kumar~Tiwari and S.~V. Nadimpalli.
\newblock Augmented random search for quadcopter control: An alternative to
  reinforcement learning.
\newblock \emph{International Journal of Information Technology and Computer
  Science}, 11\penalty0 (11):\penalty0 24–33, Nov 2019.
\newblock ISSN 2074-9015.
\newblock \doi{10.5815/ijitcs.2019.11.03}.
\newblock URL \url{http://dx.doi.org/10.5815/ijitcs.2019.11.03}.

\bibitem[Chen et~al.(2020)Chen, Wang, Haas-Heger, and
  Ciocarlie]{chen2020underactuation}
T.~Chen, L.~Wang, M.~Haas-Heger, and M.~Ciocarlie.
\newblock Underactuation design for tendon-driven hands via optimization of
  mechanically realizable manifolds in posture and torque spaces.
\newblock \emph{IEEE Transactions on Robotics}, 2020.

\bibitem[garage contributors(2019)]{garage}
T.~garage contributors.
\newblock Garage: A toolkit for reproducible reinforcement learning research.
\newblock \url{https://github.com/rlworkgroup/garage}, 2019.

\end{thebibliography}
\clearpage

\appendix

\Large \textbf{Supplementary Materials}

\normalsize

\section{Mass-spring Toy Problem - Implementation Details}
\label{appendix:toy_problem}

\subsection{Optimizing Spring Stiffnesses and Computational Policy}

\textbf{Problem Formulation.}
We have presented the problem formulation in the paper (except for the exact form of the reward function), and we list these bullet points again for better readability here.

\begin{compactitem}
    \item The \textit{observations} we can measure are the mass position $y_1$ and velocity $\dot{y}_1$.
    \item The \textit{input} to the hardware policy is motor current $i$, which is the result of the computational policy.
    \item The \textit{variables} to optimize are the weights and biases in the computational policy neural network, as well as all the spring stiffnesses $k_1, k_2, \cdots, k_n$.
    \item The \textit{goal} is to make the mass $m_2$ go to the red target line in Fig \ref{fig:toy_problem_env} and stay there, with the minimum input effort. We designed a two-stage reward function that rewards smaller position and velocity error when the mass $m_2$ is far from the goal or moving fast, and in addition rewards less input current when the mass is close to the goal and almost still.
    \begin{equation}
    \footnotesize
        R \!=\! \{\!
        \begin{array}{lr}
        \!-\alpha|y_2\!-\!h|\!-\!\beta|\dot{y}_2|\!-\!\gamma|i_{max}|,&if~\alpha|y_2\!-\!h|\!+\!\beta|\dot{y}_2|>\epsilon \\
        \!-\alpha|y_2\!-\!h|\!-\!\beta|\dot{y}_2|\!-\!\gamma|i|,&if~\alpha|y_2\!-\!h|\!+\!\beta|\dot{y}_2|<\epsilon \\
        \end{array}
    \normalsize
    \label{eq:toy_problem_reward}
    \end{equation}
    where the $\alpha$, $\beta$, and $\gamma$ are the weighting coefficients, $i_{max}$ is the upper bound of the motor current, and $\epsilon$ is a hand-tuned threshold. 
\end{compactitem}

\textbf{Shared Implementation Details.}
We implemented HWasP, HWasP-Minimal, as well as our two baselines: CMA-ES with RL inner loop, and CMA-ES. In order to have a fair comparison between them, we intentionally made different cases share common aspects wherever possible. The physics parameters not being optimized are the same for all cases: $m_1=m_2=0.1kg$, $l=0.1m$, $h=0.2m$, $g=9.8m/s^2$, $k_T=0.001Nm/A$, $r_{shaft}=0.001m$. The initial conditions are random within the feasible range. The initial values of the total spring stiffness are sampled from $0$ to $100 N/m$. We used PPO and CMA-ES in the Garage package \cite{garage} for all cases. We implemented the computational graphs for HWasP in TensorFlow and the dynamics of the rest of the environment (non-differentiable) by ourselves using mid-point Euler integration. In the computational policies the neural network sizes are set to be 2 layers and 32 nodes each layer. The episode length is $1,000$ environment steps and the total number of steps is $4 \times 10^6$.

\textbf{Hardware as Policy.}
We use Hooke's Law for the parallel springs
\begin{equation}
    f_{spr} = \sum_{i=1}^{n}k_iy_1
\end{equation}
and current-torque relationship for the motor (ignoring rotor inertia and friction)
\begin{equation}
    f_{str} = \tfrac{k_Ti}{r_{shaft}}
\end{equation}
to model the mechanical part of our agent. We implement the computational graph of this hardware policy and combine it with a computational policy expressed as a neural network computational, as shown in Fig \ref{subfig:toy_problem_hw_as_policy_comp_graph}.

\textbf{Hardware as Policy --- Minimal.}
In this case 
(Fig.~\ref{subfig:toy_problem_hw_as_action_comp_graph}),
we only add the hardware parameters (spring stiffnesses vector $\bm{k}$) to the action vector of the computational policy. The environment is governed by the physics of the spring-mass system, and takes the new values of $\bm{k}$ into account when simulating of the next time step. 

\setcounter{figure}{5}    

\begin{figure}[t!]
    \begin{subfigure}[b]{0.49\linewidth}
        \centering
        \includegraphics[width=0.7\linewidth]{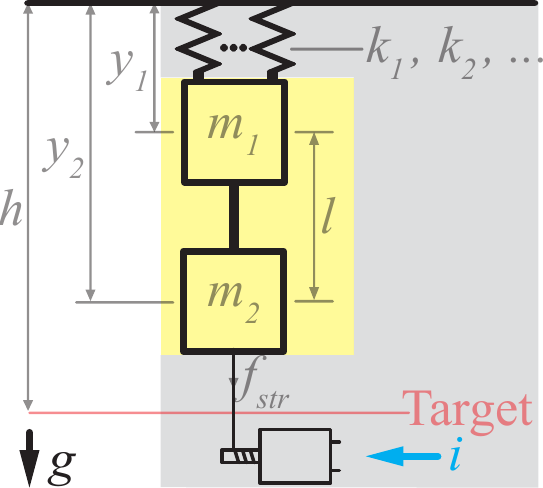}
        \caption{HWasP}
        \label{subfig:toy_problem_hw_as_policy_env}
    \end{subfigure}
    \begin{subfigure}[b]{0.49\linewidth}
        \centering
        \includegraphics[width=0.7\linewidth]{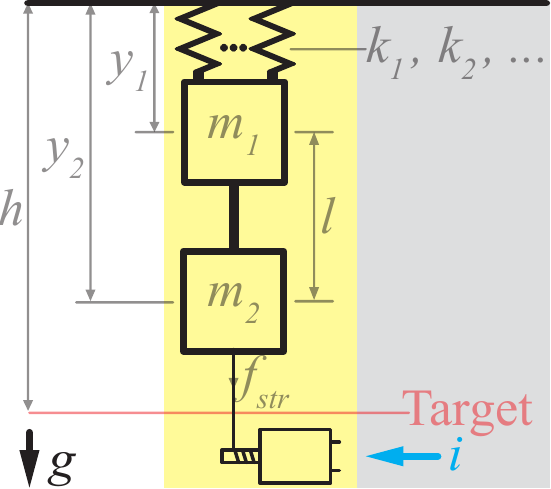}
        \caption{HWasP-Minimal}
        \label{subfig:toy_problem_hw_as_action_env}
    \end{subfigure}
    \caption{The mass-spring system --- optimizing spring stiffnesses and computational policy (yellow: environment, gray: agent/policy).  (This is already presented in the paper, and we include it again for better readability.)}
    \label{fig:toy_problem_env}
\end{figure}

\begin{figure}[t!]
    \begin{subfigure}[b]{0.49\linewidth}
        \centering
        \includegraphics[width=0.85\linewidth]{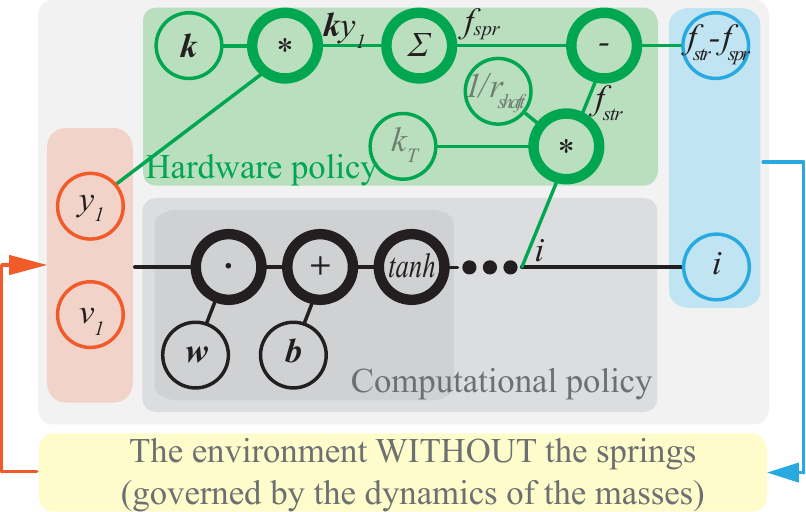}
        \caption{HWasP}
        \label{subfig:toy_problem_hw_as_policy_comp_graph}
    \end{subfigure}
    \begin{subfigure}[b]{0.49\linewidth}
        \centering
        \includegraphics[width=0.85\linewidth]{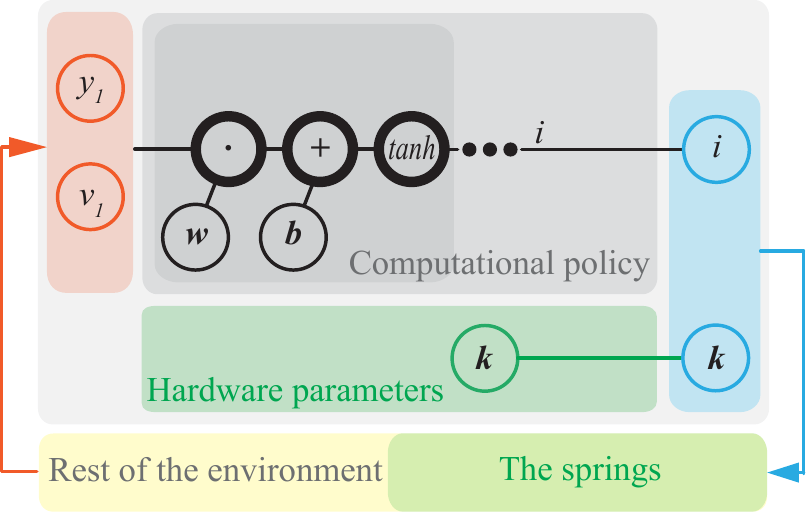}
        \caption{HWasP-Minimal}
        \label{subfig:toy_problem_hw_as_action_comp_graph}
    \end{subfigure}
    \caption{The (simplified) computational graphs of the proposed method for the toy problem --- optimizing spring stiffnesses and computational policy. Bold circles represent tensor operations, light black or green circles with black fonts represent the variables to be optimized, and light circles with translucent fonts represent constants that are not part of the optimization.}
    \label{fig:toy_problem_comp_graphs}
\end{figure}

\textbf{Numerical Results.} If we ignore the transition phase of the task in this toy problem and make a quasi-static assumption, and assuming total spring stiffness equals the following:
\begin{equation}
    k^* = \tfrac{(m_1+m_2)g}{h-l}
\label{eq:toy_problem_optimal_k}
\end{equation}
then gravity will drag the mass $m_2$ exactly to the target, and the steady-state input current $i$ can be zero, which minimizes the return on $i$ in a long enough horizon. In the real world, the system has dynamic effects, but optimized total stiffness $k$ should still be close to this value given a long enough horizon. After training, we indeed find the optimized total stiffness close to  $k^*$, as shown in Table \ref{tab:toy_problem_result}.

\begin{table}[h!]\centering
\caption{Optimization results of total stiffness [N/m]}
\label{tab:toy_problem_result}
\begin{tabular}{ c|cc } 
                        & $n=10$ & $n=50$ \\ 
    \hline
    $k^*$ & \multicolumn{2}{c}{19.6} \\
    HWasP & 19.9 & 20.1 \\ 
    HWasP-Minimal & 18.7 & 21.9 \\ 

\end{tabular}
\end{table}

\subsection{Optimizing Bar Length and Computational Policy}
\begin{figure}[t!]
    \begin{subfigure}[b]{0.49\linewidth}
        \centering
        \includegraphics[width=0.7\linewidth]{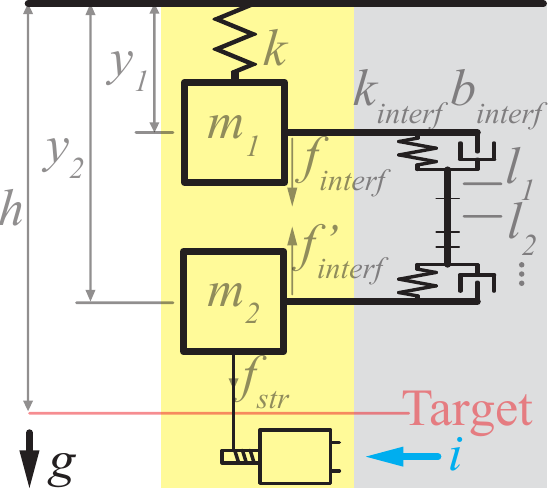}
        \caption{HWasP}
        \label{subfig:toy_problem_opt_l_hw_as_policy_env}
    \end{subfigure}
    \begin{subfigure}[b]{0.49\linewidth}
        \centering
        \includegraphics[width=0.7\linewidth]{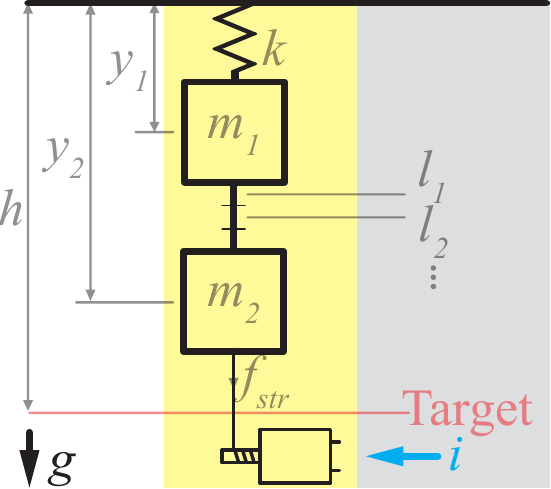}
        \caption{HWasP-Minimal}
        \label{subfig:toy_problem_opt_l_hw_as_action_env}
    \end{subfigure}
    \caption{The mass-spring system --- optimizing the bar lengths and computational policy (yellow: environment, gray: agent/policy). }
    \label{fig:toy_problem_opt_l_env}
\end{figure}

Another case we tried for the toy problem is to optimize the bar length connecting two masses. Unlike the previous case, the interface between the redefined policy and environment is no longer an element generating forces, but a ``hard'' geometric relationship. For HWasP, we propose to use springs and dampers to ``soften'' such interfaces, and still use the spring-damper forces as the redefined action. For HWasP-Minimal, the bar length can directly be used a part of the action vector. Similar to the previous case, we divide the bar into multiple segments and treat the length of each segment as an individual parameter to test the scalability to higher-dimensional problems. The spring-mass system is illustrated in Fig.\ref{fig:toy_problem_opt_l_env}. 

\begin{figure}[t!]
    \begin{subfigure}[b]{0.49\linewidth}
        \centering
        \includegraphics[width=0.95\linewidth]{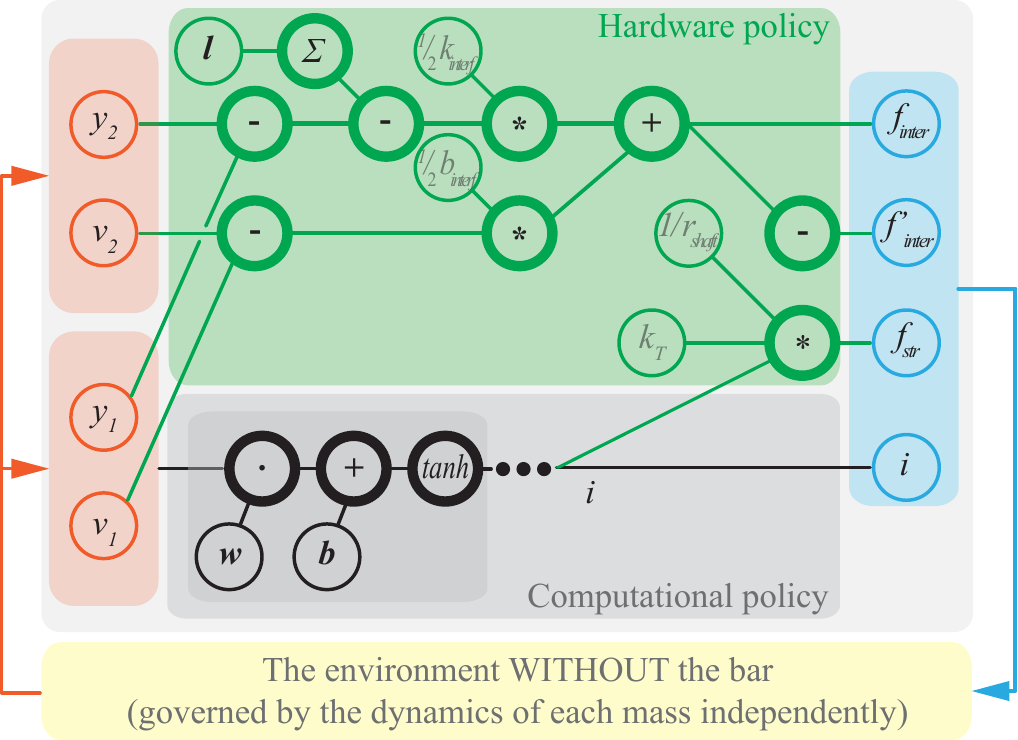}
        \caption{HWasP}
        \label{subfig:toy_problem_opt_l_hw_as_policy_comp_graph}
    \end{subfigure}
    \begin{subfigure}[b]{0.49\linewidth}
        \centering
        \includegraphics[width=0.95\linewidth]{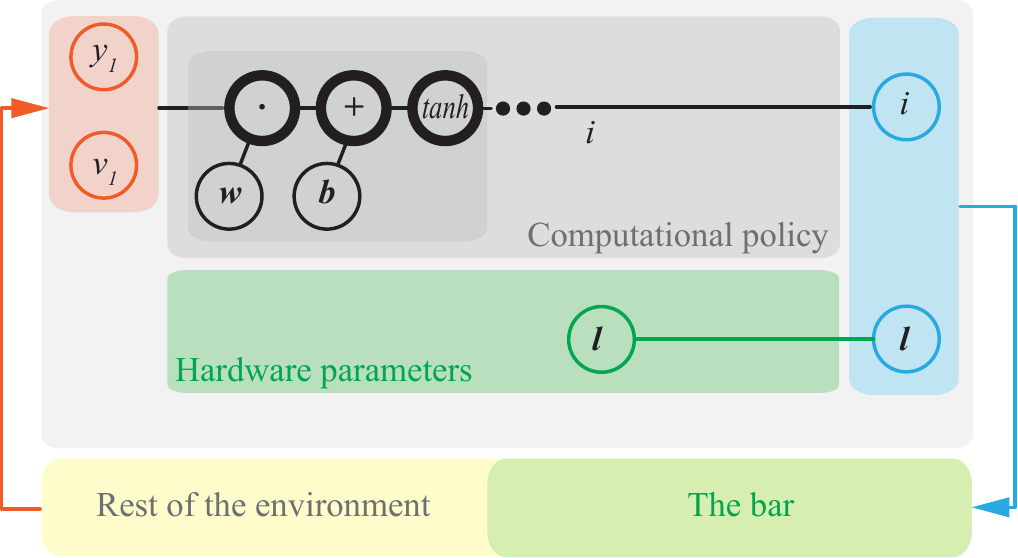}
        \caption{HWasP-Minimal}
        \label{subfig:toy_problem_opt_l_hw_as_action_comp_graph}
    \end{subfigure}
    \caption{The (simplified) computational graphs of the proposed method for the mass-spring toy problem --- optimizing bar lengths and computational policy.}
    \label{fig:toy_problem_opt_l_comp_graphs}
\end{figure}

\begin{figure}[h!]
    \centering
    \includegraphics[width=0.6\linewidth]{images/toy_problem_learning_curve_legend.pdf}
    \begin{subfigure}[b]{0.49\linewidth}
        \centering
        \includegraphics[width=0.95\linewidth]{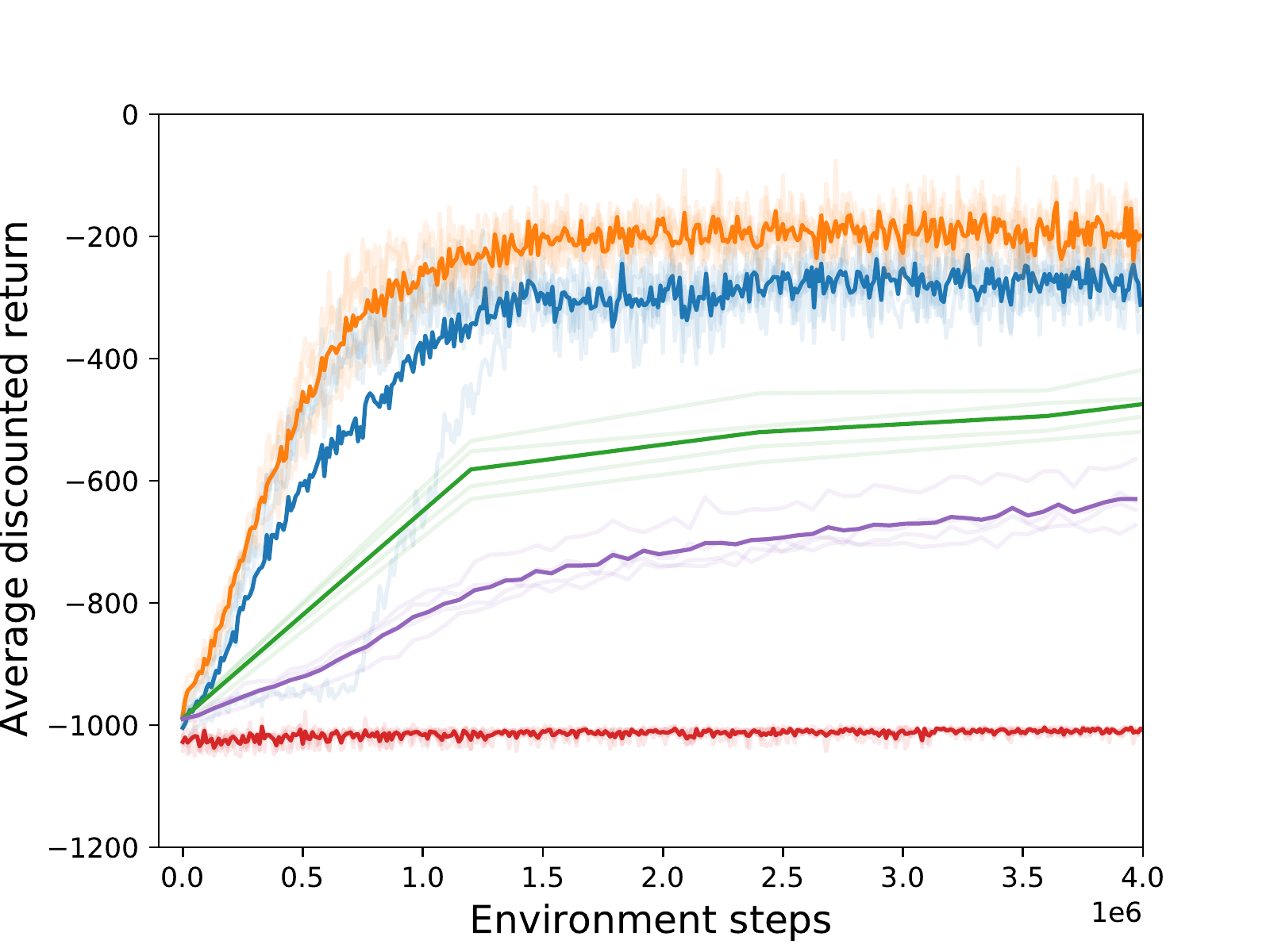}
        \caption{10 parameters}
        \label{subfig:toy_problem_opt_l_hw_as_policy_training_curves_10_params}
    \end{subfigure}
    \begin{subfigure}[b]{0.49\linewidth}
        \centering
        \includegraphics[width=0.95\linewidth]{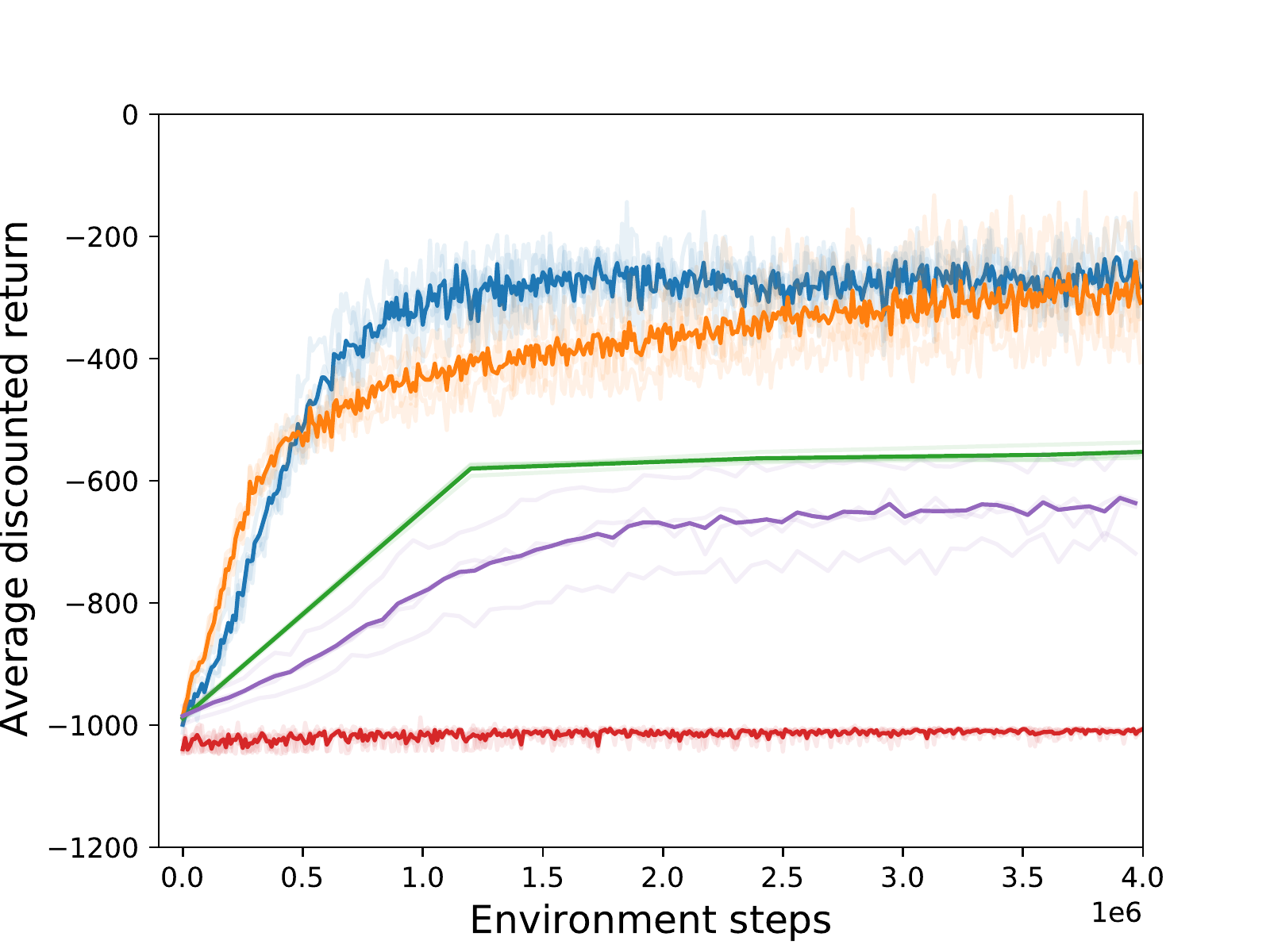}
        \caption{50 parameters}
        \label{subfig:toy_problem_opt_l_hw_as_policy_training_curves_50_params}
    \end{subfigure}
    \caption{The training curves of the toy problem --- optimizing bar lengths and computational policy.}
    \label{fig:toy_problem_opt_l_training_curves}
\end{figure}

Since we assume the bar is weightless, the dynamics is infinitely fast. Assuming all interfaces have the same $k_{interf}$ and $b_{interf}$, we have:

\begin{equation} 
\label{f_interf}
\begin{split}
  f_{interf} & = k_{interf} ((\frac{1}{2}(y_1 + y_2) - \frac{1}{2}\sum_{i=1}^{n}l_i) - y_1)) + b_{interf}(\frac{1}{2}(\dot{y}_1+\dot{y}_2) - \dot{y}_1) \\
   & =\frac{1}{2}k_{interf}(y_2 - y_1 - \sum_{i=1}^{n}l_i) + \frac{1}{2}b_{interf}(\dot{y}_2 - \dot{y}_1), \\
\end{split}
\end{equation}

\begin{equation}
    f'_{interf} = -f_{interf}.
\end{equation}
We implement this relationship as a computational graph, as shown in Fig.\ref{subfig:toy_problem_opt_l_hw_as_policy_comp_graph}.

Except for the variables to optimize and the computational graphs, other aspects (such as the reward function) are the same as the previous case. The training curves are shown in Fig.\ref{fig:toy_problem_opt_l_training_curves}. We can see that the performance of both our methods and the baselines are very similar to the previous case.

\section{Co-Design of an Underactuated Hand - Implementation Details}
\label{appendix:hand}
\textbf{Tendon Underactuation Model.}
In a tendon-driven underactuated hand, the tendon mechanism acts as the transmission that converts motor forces to joint torques, based on the overall state of the system. In the model we constructed, it assumes an elastic tendon (with stiffness $k_{tend}$) goes through multiple revolute joints by wrapping around circular pulleys (radii $\bm{r}_{pul}$), and each joint is closed by the tendon and opened by a restoring spring (stiffness $\bm{k}_{spr}$ and preload angle $\bm{\theta}_{spr}^{pre}$). In order to make this problem determinate, our model thus assumes a nominal (finite) tendon stiffness, takes in the commanded relative motor travel $\Delta x_{mot}$, the motor position reading $x_{mot}$ and the joint angles $\bm{\theta}_{joint}$ in the previous time step, and computes the joint torques $\bm{\tau}_{joint}$ for the current time step. Such torques are then commanded to the joints in the physics simulation. 

The tendon elongation can be calculated as:
\begin{equation}
    \delta_{tend} = (x_{mot} + \Delta x_{mot}) + \bm{r}_{pul}^T \bm{\theta}_{joint}^{ref} - \bm{r}_{pul}^T \bm{\theta}_{joint}
    \label{eq:ua_model_elongation}
\end{equation}
where $\bm{\theta}_{joint}^{ref}$ is the joint angle when the motor is in zero-position, and usually we define it to be zero. Then the tendon force can be calculated as:
\begin{equation}
    f_{tend} = k_{tend} \delta_{tend} .
    \label{eq:ua_model_force}
\end{equation}
Hence, the torques applied to joints are:
\begin{equation}
    \bm{\tau}_{joint} = f_{tend} \bm{r}_{pul} - \bm{k}_{spr} \bm{*} (\bm{\theta}_{joint} + \bm{\theta}_{spr}^{pre})
    \label{eq:ua_model_trq}
\end{equation}
where $\bm{*}$ means element-wise multiplication.

\begin{figure}[t]
    \begin{subfigure}{0.49\linewidth}
        \centering
        \includegraphics[width=0.95\linewidth]{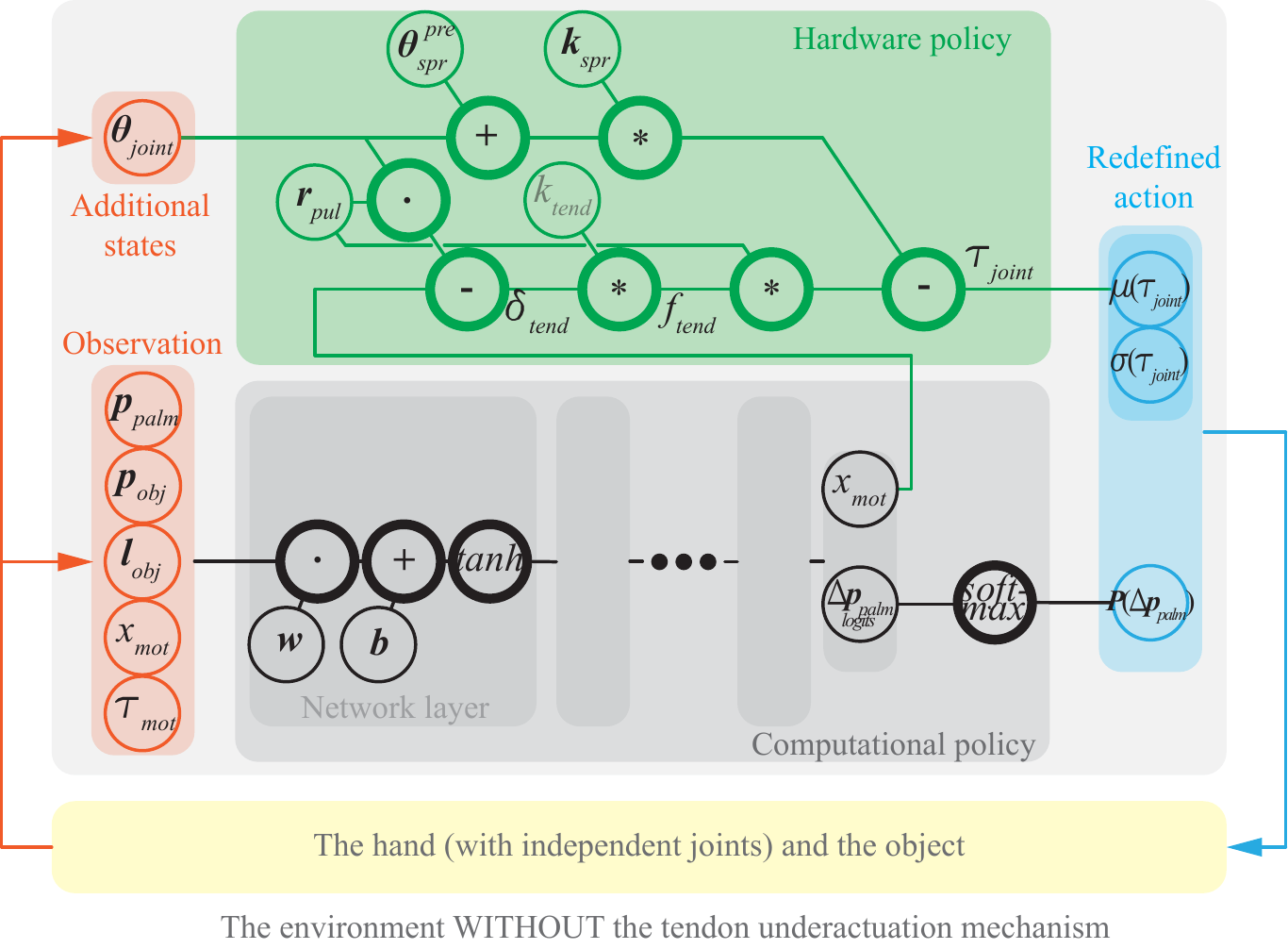}
        \caption{HWasP}
        \label{subfig:nasa_hand_hw_as_policy_comp_graph}
    \end{subfigure}
    \begin{subfigure}{0.49\linewidth}
        \centering
        \includegraphics[width=0.95\linewidth]{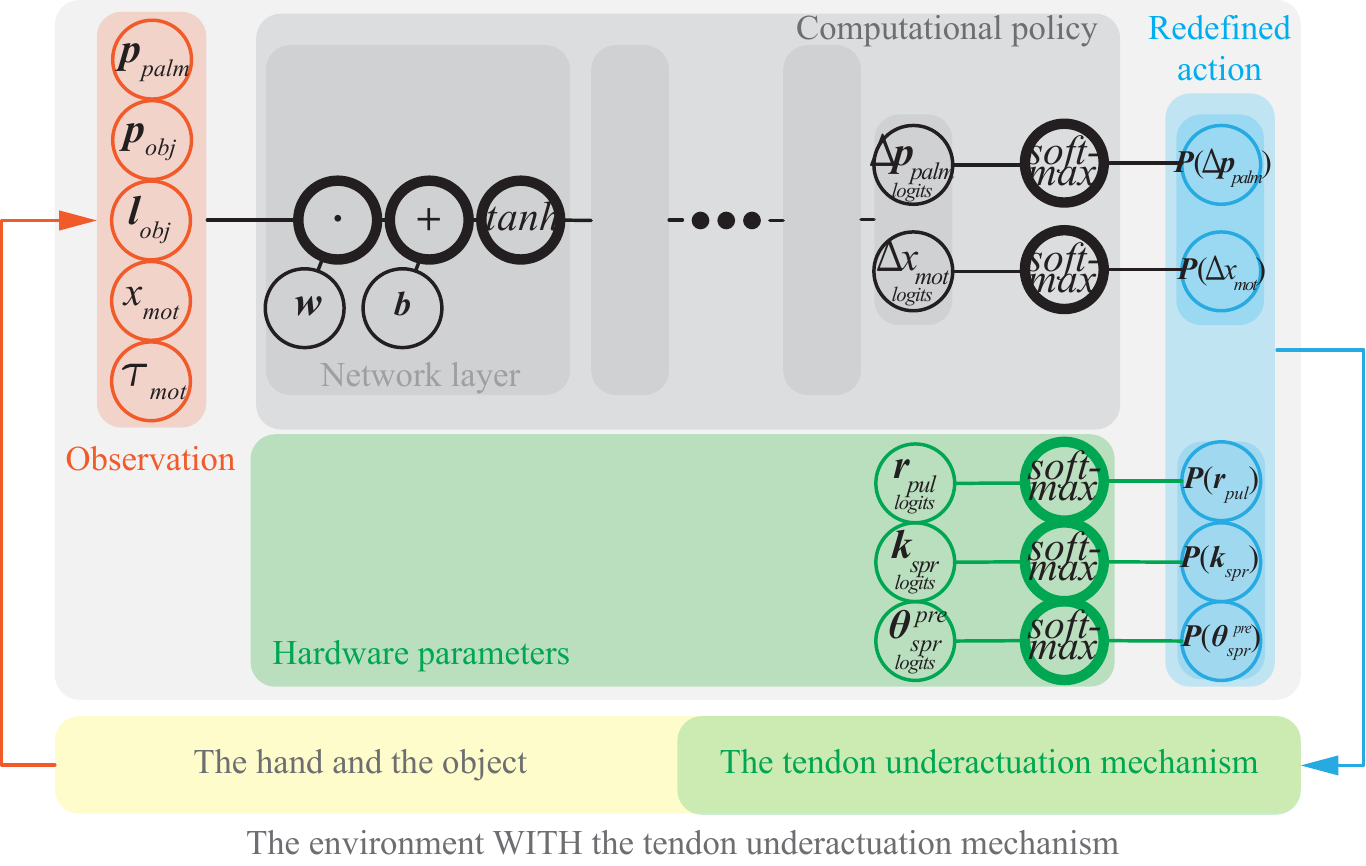}
        \caption{HWasP-Minimal}
        \label{subfig:nasa_hand_hw_as_action_comp_graph}
    \end{subfigure}
    \caption{The (simplified) computational graphs of both proposed methods for the hand mechanical-computational co-design problem. The plotting conventions are the same as the toy problem.}
    \label{fig:nasa_hand_comp_graph}
\end{figure}

This model is built as an auto-differentiable computational graph in HWasP, and a non-differentiable model on top of the physics simulation in HWasP-Minimal and the baselines.

\textbf{Problem Formulation.}

\begin{compactitem}
    \item The \textit{observations} are the position vector of the palm $\bm{p}_{palm}$, the position vector of the object $\bm{p}_{obj}$, the size vector of the object bounding-box $\bm{l}_{obj}$, and the current hand motor travel $x_{mot}$ and torque $\tau_{mot}$. We can also send joint angles $\bm{\theta}_{joint}$ to the hardware policy (only used in training time), but not the computational policy, because there are no joint encoders in such an underactuated hand, and the computational policy (which will serve as a controller of the real robot in run time) does not have access to joint angles.
    
    \item The \textit{input} to the hardware policy also includes the relative motor travel command $\Delta x_{mot}$ produced by the computational policy. The output of the overall system includes joint torques as well as palm motion $\Delta \bm{p}_{palm}$.
    
    \item The \textit{variables} to optimize are the parameters in the computational policy neural network, and the hand underactuation parameters: pulley radii $\bm{r}_{pul}$, the joint restoring spring stiffnesses $\bm{k}_{spr}$ and the joint restoring spring preload angles $\bm{\theta}_{spr}^{pre}$, where each vector has a dimension of four corresponding to the proximal and distal joints in the thumb and the opposing fingers (the two fingers share the same parameters). 
    
    \item The \textit{goal} is to grasp the object and lift it up. Formally, the reward function is:
    \begin{equation}
    R = \alpha \left\lVert \bm{p}_{palm} - \bm{p}_{obj}\right\rVert + \beta C + f(z_{obj})
    \label{eq:nasa_hand_reward}
    \end{equation}
    where the $\alpha$, and $\beta$ are the weighting coefficients, $\bm{p}_{palm}$ and $\bm{p}_{obj}$ are the positions of the palm and the object, $C$ is the number of contacts between the distal links and the object, $z_{obj}$ is the height of the object, and $f(z_{obj})$ is a hand-tuned non-decreasing piecewise-constant function of $z_{obj}$.
\end{compactitem}

\textbf{Shared Implementation Details.}
Similar to the toy problem, we used the Garage package \cite{garage} and TensorFlow for HWasP, HWasP-Minimal and the two baselines in this design case. We use MuJoCo \cite{todorov2012mujoco} for the physics simulation of the hand. The simulation time step is $0.001s$, the environment step is $0.01s$, and there are 500 environment steps per episode and $2 \times 10^7$ steps for the entire training.
The initial height of the hand, as well as the type, weight, size, and friction coefficient of the object are randomly sampled within a reasonable range.
We also added random perturbations forces and torques on the hand-object system to encourage more stable grasps. We use TRPO to explicitly limit the shift of action distributions.
The computational policy is a fully-connected neural network with 2 layers and 128 hidden units on each layer. 
To stabilize the training, we scale the parameters in the hardware model by changing their units. Specifically, we scale all tendon stiffness and $k_{spr}$ by $1.0e^{-3}$. Then, we multiply the torques by $1.0e^3$ when we apply them to the environment.

\textbf{Hardware as Policy.}
Shown in Fig. \ref{subfig:nasa_hand_hw_as_policy_comp_graph}, we implement the underactuation model as a hardware policy computational graph, which allows the hardware parameters to be optimized via auto-differentiation and back-propagation. We redefine the action in the RL formulation to be the relative palm position command as well as the joint torques. The environment then becomes the hand-object system without the tendon underactuation mechanisms, i.e. with independent joints.

\textbf{Hardware as Policy --- Minimal.}
We also implemented the ``HWasP-Minimal'' method by incorporating all hardware parameters (pulley radii $\bm{r}_{pul}$, joint restoring spring stiffnesses $\bm{k}_{spr}$ and preload angle $\bm{\theta}_{spr}^{pre}$) into the original control (relative hand position $\Delta \bm{p}_{palm}$ and motor command $\Delta x_{mot}$), shown in Fig. \ref{subfig:nasa_hand_hw_as_action_comp_graph}. The environment (containing underactuated hand and the object) takes in all these actions, sets the hardware parameters to the tendon model and performs simulation.

\textbf{Numerical Results.}
Our results show that we can learn effective hardware parameters. The resulting pulley radii, spring stiffnesses and preload angles are shown in Table \ref{tab:nasa_hand_resulted_params}, using HWasP and HWasP-Minimal respectively. We note that the resulting parameters do not necessarily need to be identical: the optimal set of underactuation parameters is not unique by nature (for example, scaling them does not change the grasping behavior; for another example, a higher spring stiffness and a higher spring preload have similar effects), the evaluation is also noisy since we intentionally injected noise, and the gradient-based training process may also settle in local optima in the optimization landscape.

\vspace{-2mm}
\begin{table}[H]
\centering
\caption{The optimized pulley radii, joint spring stiffnesses and preload angles. In each cell, the first number is from HWasP, and the second number is from HWasP-Minimal.}
\label{tab:nasa_hand_resulted_params}
\begin{tabular}{ c|c|c|c } 
    & Pulley radius {[}mm{]} & Spring stiffness {[}Nmm/rad{]} & Spring preload {[}rad{]} \\
    \hline
        Thumb proximal & 10.0~~~~~~8.1 & 6.2~~~~~~~~7.0 & 2.0~~~~~~~~3.2 \\
        Thumb distal & 7.5~~~~~~~~6.1 & 6.1~~~~~~~~9.2 & 1.5~~~~~~~~3.1 \\
        Finger proximal & 3.0~~~~~~~~5.0 & 6.1~~~~~~~~8.2 & 1.9~~~~~~~~3.4 \\ 
        Finger distal & 2.6~~~~~~~~4.0 & 5.9~~~~~~~~9.2 & 1.1~~~~~~~~3.1 \\ 
\end{tabular}
\end{table}

\begin{figure}[t!]
    \begin{subfigure}[b]{0.32\linewidth}
        \centering
        \includegraphics[width=\linewidth]{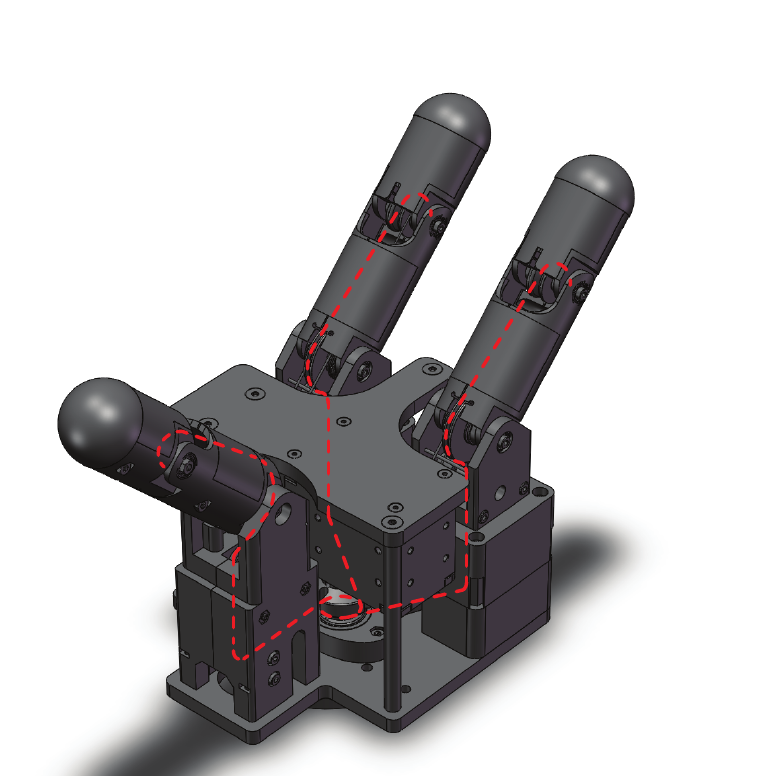}
        \caption{The CAD design and tendon \\ routing (red lines).}
        \label{subfig:nasa_hand_cad}
    \end{subfigure}
    \begin{subfigure}[b]{0.3\linewidth}
        \centering
        \includegraphics[width=0.7\linewidth]{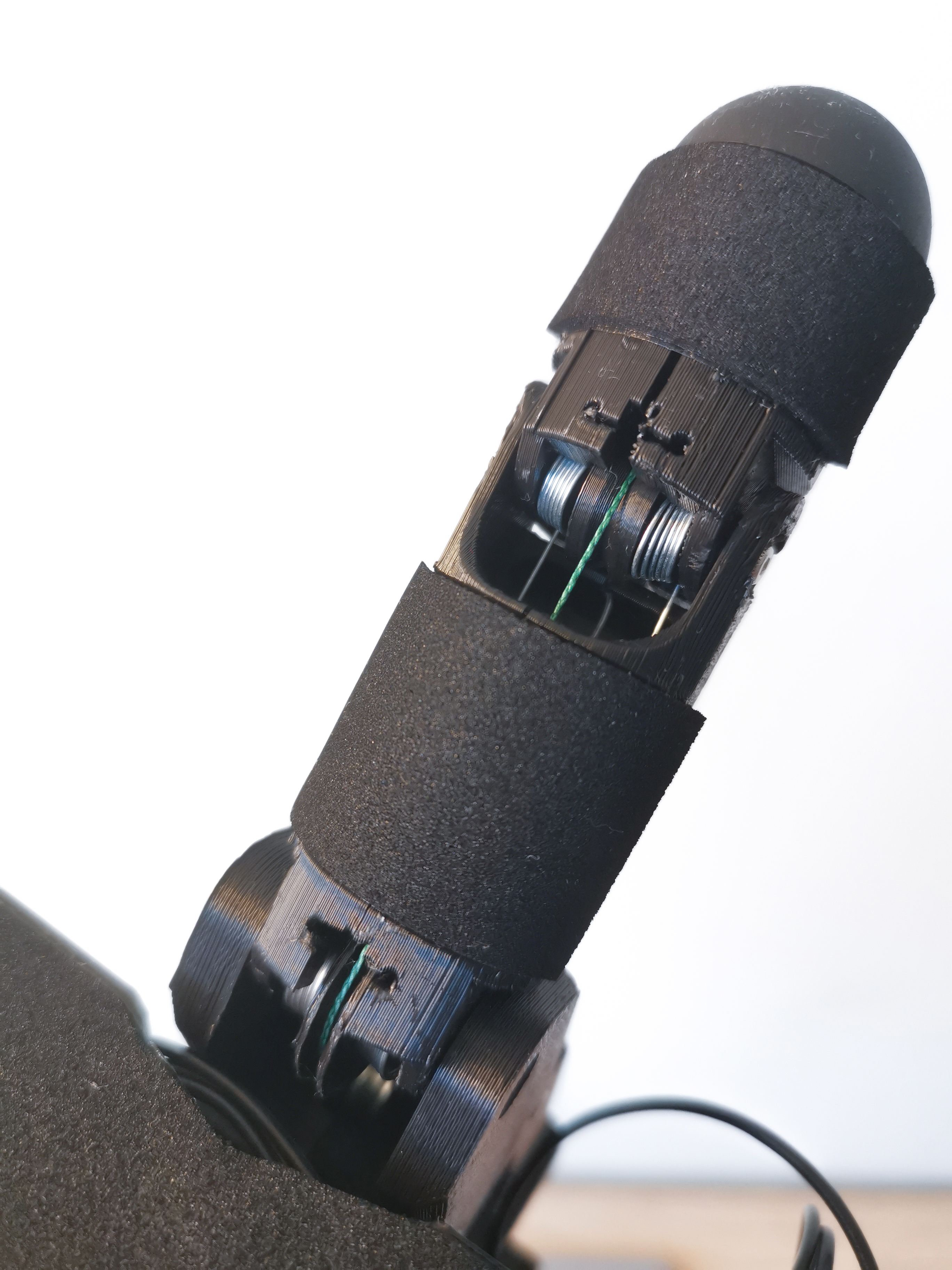}
        \caption{A close-up showing the \\ joints, tendons and springs.}
        \label{subfig:nasa_hand_tendon}
    \end{subfigure}
    \begin{subfigure}[b]{0.35\linewidth}
        \centering
        \begin{tabular}{cc}
        \includegraphics[width=0.5\linewidth]{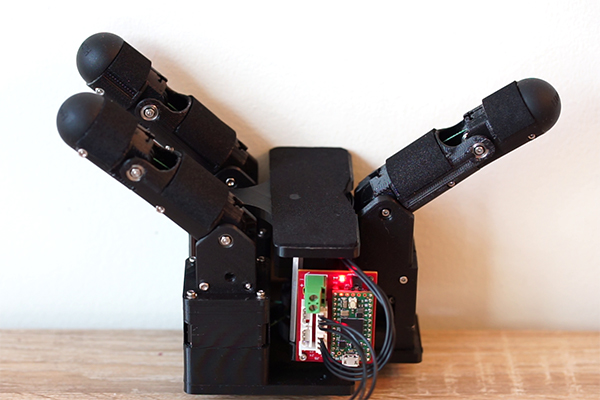} \hspace{-5mm}
        & \includegraphics[width=0.5\linewidth]{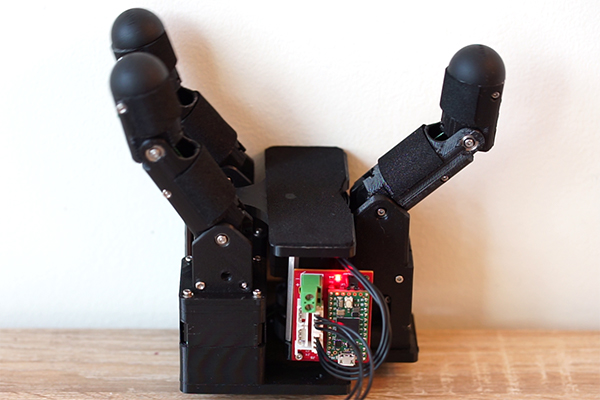} \\
        \includegraphics[width=0.5\linewidth]{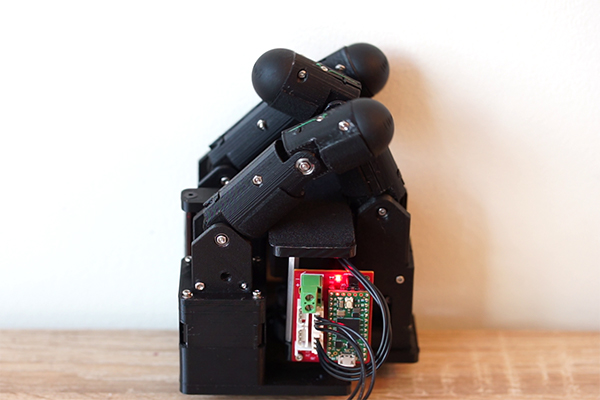} \hspace{-5mm}
        & \includegraphics[width=0.5\linewidth]{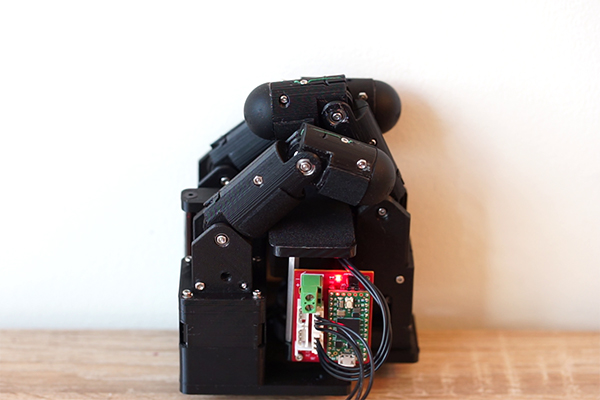} \\
        \end{tabular}
        \centering
        \caption{The finger trajectory. \newline }
        \label{subfig:nasa_hand_tendon}
    \end{subfigure}
    
    \caption{The physical hand prototype.}
    \label{fig:nasa_hand_prototype}
\end{figure}

\textbf{Validation with Physical Prototype.}
Here we present additional details of the physical prototype we built according to optimization results. This hand is 3D printed using polylactide (PLA). All eight joints in three fingers are actuated by a single servo motor (DYNAMIXEL XM430-W210-T) equipped with 12-bit encoder and current sensing. The tendons are made of Ultra-high-molecular-weight polyethylene (commercially named Spectra$^{\tiny{\textregistered}}$). In each finger joint, there is a pulley with the radius prescribed by the optimization, and two parallel springs whose stiffnesses also add up to the optimized value prescribed by our method. The distal joint pulleys are fixed to the finger structures, and proximal joint pulleys are free-rotating. 
The CAD model, tendon routing scheme, joint design, and finger trajectory are shown in Fig.  \ref{fig:nasa_hand_prototype}.

\textbf{Sim-to-Real Transfer.}
In order to transfer the trained policy from the simulation to the physical world, we used the following techniques:
\begin{compactitem}
    \item We applied the \textit{Domain Randomization} techniques to a lot of physical parameters and processes. We randomized object shape (among sphere, box, cylinder, ellipsoid), size (bounding box size uniformly sampled from $40$ to $100mm$), weight (uniformly sampled from $100$ to $500g$), friction coefficient (uniformly sampled from $0.5$ to $1.0$) and inertia (each principal component uniformly sampled from $0.0001$ to $0.005 kg \cdot m^2$). We also injected sensor and actuation noise (Gaussian noise with $1mm$ and $0.01rad$ standard deviation for translational and rotational joints respectively), and applied random disturbance wrenches (Gaussian disturbance with $0.02N$ and $0.002Nm$ standard deviation for force and torque respectively) on the hand-object system.
    \item In simulation, we limited the hand motion close to quasi-static, and use position control to drive the palm and hand joints. This control scheme is not sensitive to inaccurate parameters, unmodeled dynamics and can effectively reject disturbances.
\end{compactitem}

\begin{figure}[t!]
    \begin{subfigure}[b]{0.32\linewidth}
        \centering
        \includegraphics[width=\linewidth]{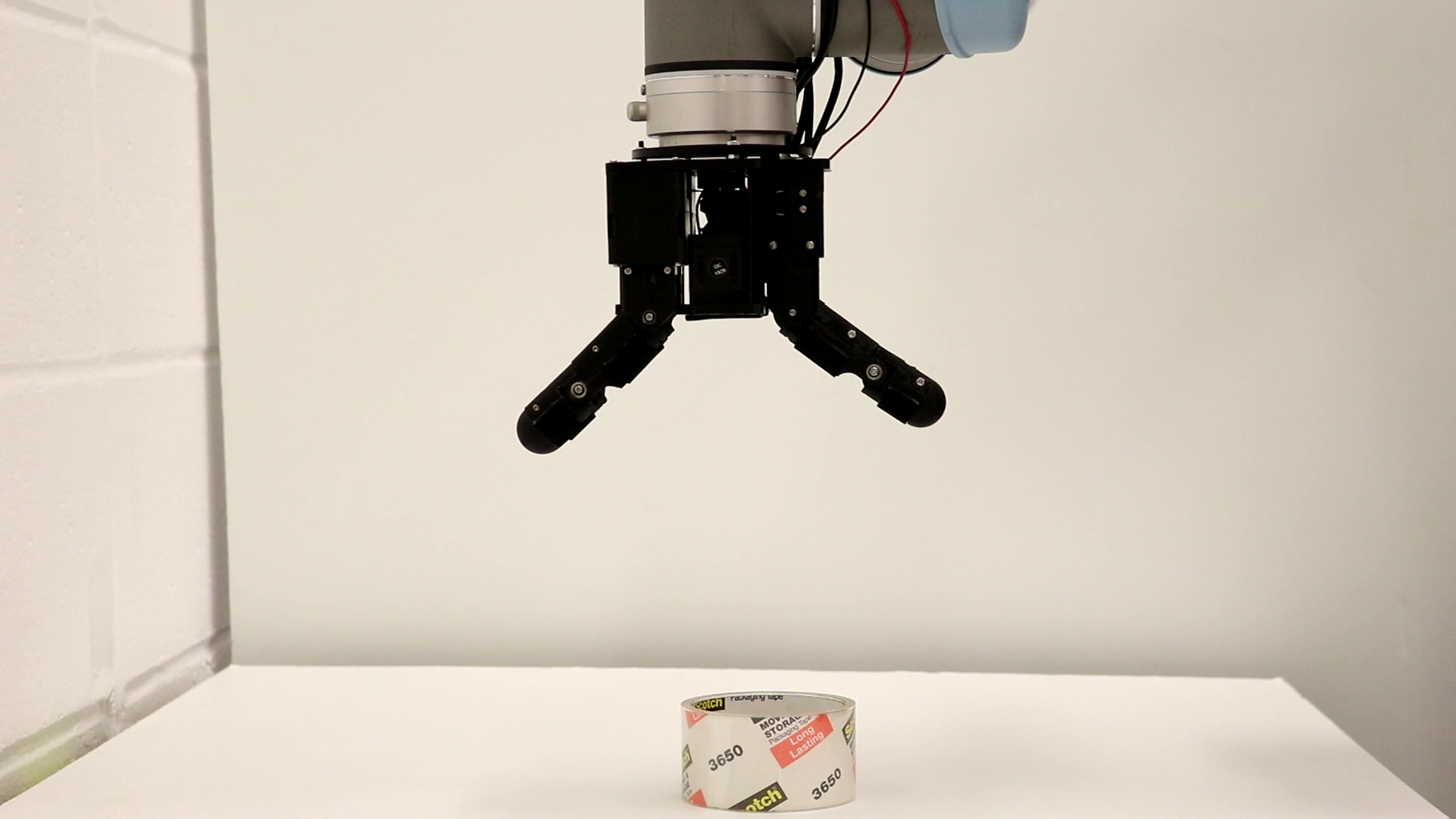}
        \label{subfig:real_robot_1}
    \end{subfigure}
    \begin{subfigure}[b]{0.32\linewidth}
        \centering
        \includegraphics[width=\linewidth]{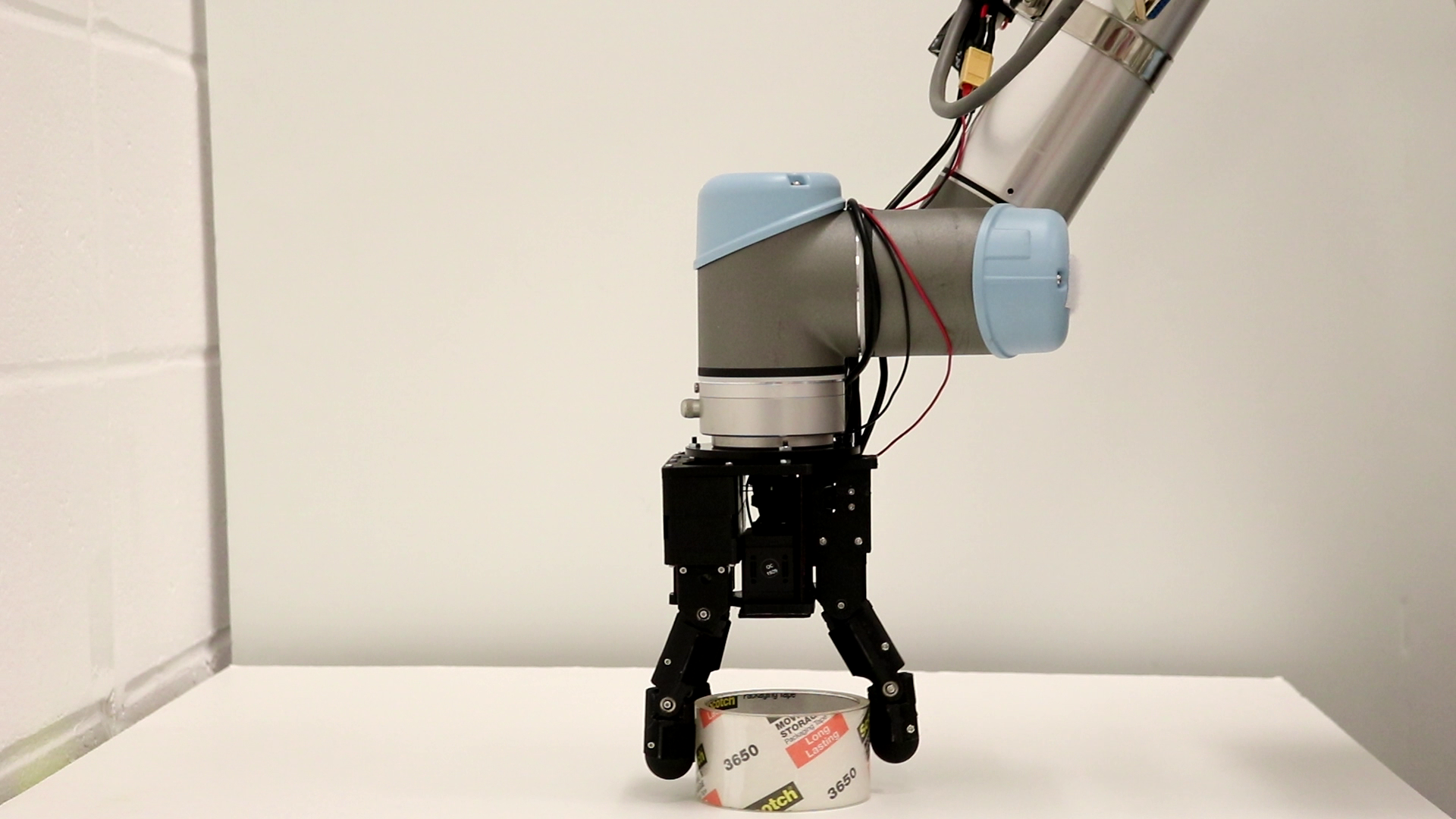}
        \label{subfig:real_robot_2}
    \end{subfigure}
    \begin{subfigure}[b]{0.32\linewidth}
        \centering
        \includegraphics[width=\linewidth]{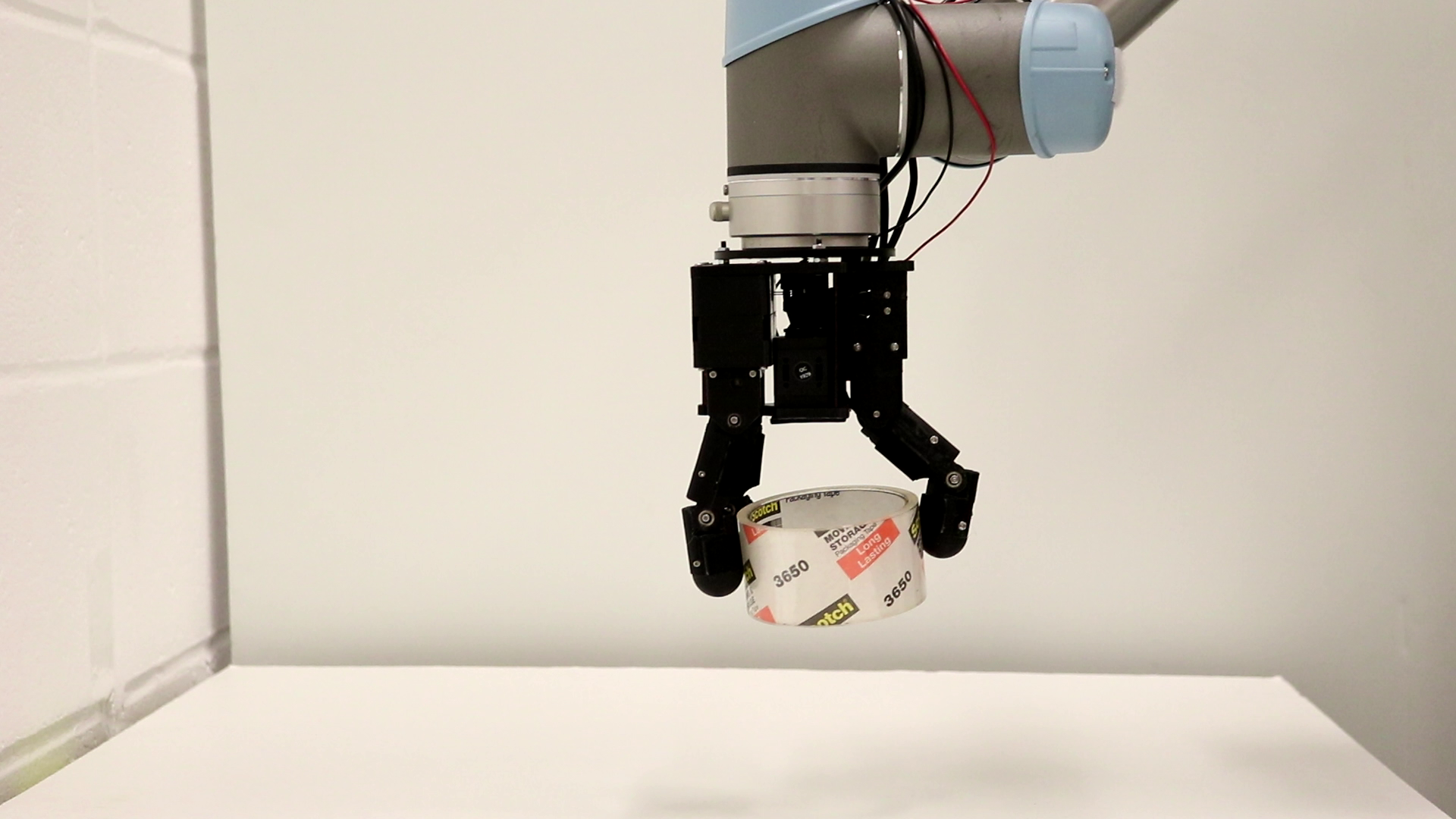}
        \label{subfig:real_robot_3}
    \end{subfigure}
    
    \caption{Policy deployment on real robot.}
    \label{fig:policy_deployment}
\end{figure}

\textbf{Policy Deployment on Real Robot.}
We used a UR5 robot and our optimized hand prototype as the hardware platform and deployed the trained computational policy to control the arm and hand motion. Fig. \ref{fig:policy_deployment} shows an example grasp in the 3D-Grasp case. Videos are available at \url{https://roamlab.github.io/hwasp/} .

\end{document}